%% file: main.tex
\documentclass{article}

\PassOptionsToPackage{numbers, compress}{natbib}
\usepackage[final]{styles/neurips_2021}

\usepackage[utf8]{inputenc} 
\usepackage[T1]{fontenc}    
\usepackage{hyperref}       
\usepackage{url}            
\usepackage{booktabs}       
\usepackage{amsfonts}       
\usepackage{nicefrac}       
\usepackage{microtype}      
\usepackage[bottom]{footmisc}

\usepackage{amsmath, amsthm, amssymb, pifont, subfig, csquotes, dashrule, tikz, bbm, booktabs, bm, graphicx, microtype, multirow, verbatim, url, mathrsfs, algorithm, algpseudocode, mdframed, enumitem}
\usepackage{hyperref}
\usepackage[capitalize,nameinlink]{cleveref}

\usepackage[colorinlistoftodos,prependcaption]{todonotes}

\input{styles/neurips_commands}
\newcommand{\taskq}{\textsc{TaskQ}}
\newcommand{\exprq}{\textsc{ExpressionQ}}

\title{On the Expressivity of Markov Reward}

\author{%
  David Abel\\
  DeepMind\\
  \texttt{dmabel@deepmind.com} \\
  \And
  Will Dabney \\
  DeepMind \\
  \texttt{wdabney@deepmind.com} \\
  \And
  Anna Harutyunyan \\
  DeepMind \\
  \texttt{harutyunyan@deepmind.com} \\
  \And
  Mark K. Ho \\
  Department of Computer Science \\
  Princeton University \\
  \texttt{mho@princeton.edu} \\
  \And
  Michael L. Littman \\
  Department of Computer Science \\
  Brown University \\
  \texttt{mlittman@cs.brown.edu}\\
  \AND
  Doina Precup \\
  DeepMind \\
  \texttt{doinap@deepmind.com}\\
  \And 
  Satinder Singh \\
  DeepMind \\
  \texttt{baveja@deepmind.com} \\
}

\begin{document}
\maketitle

\begin{abstract}
%
Reward is the driving force for reinforcement-learning agents.
%
This paper is dedicated to understanding the expressivity of reward as a way to capture tasks that we would want an agent to perform.
%
We frame this study around three new abstract notions of ``task'' that might be desirable: (1) a set of acceptable behaviors, (2) a partial ordering over behaviors, or (3) a partial ordering over trajectories.
%
Our main results prove that while reward can express many of these tasks, there exist instances of each task type that no Markov reward function can capture.
%
We then provide a set of polynomial-time algorithms that construct a Markov reward function that allows an agent to optimize tasks of each of these three types, and correctly determine when no such reward function exists.
%
We conclude with an empirical study that corroborates and illustrates our theoretical findings.
\end{abstract}

\section{Introduction}

%
How are we to use algorithms for reinforcement learning (RL) to solve problems of relevance in the world? Reward plays a significant role as a general purpose signal: For any desired behavior, task, or other characteristic of agency, there must exist a reward signal that can incentivize an agent to learn to realize these desires. Indeed, the expressivity of reward is taken as a backdrop assumption that frames RL, sometimes called the reward hypothesis: ``...all of what we mean by goals and purposes can be well thought of as maximization of the expected value of the cumulative sum of a received scalar signal (reward)''~\cite{suttonwebRLhypothesis,littmanRH,christian2021alignment}. In this paper, we establish first steps toward a systematic study of the reward hypothesis by examining the expressivity of reward as a signal. We proceed in three steps.

\begin{figure}[t!]
    \centering
    \includegraphics[width=0.65\textwidth]{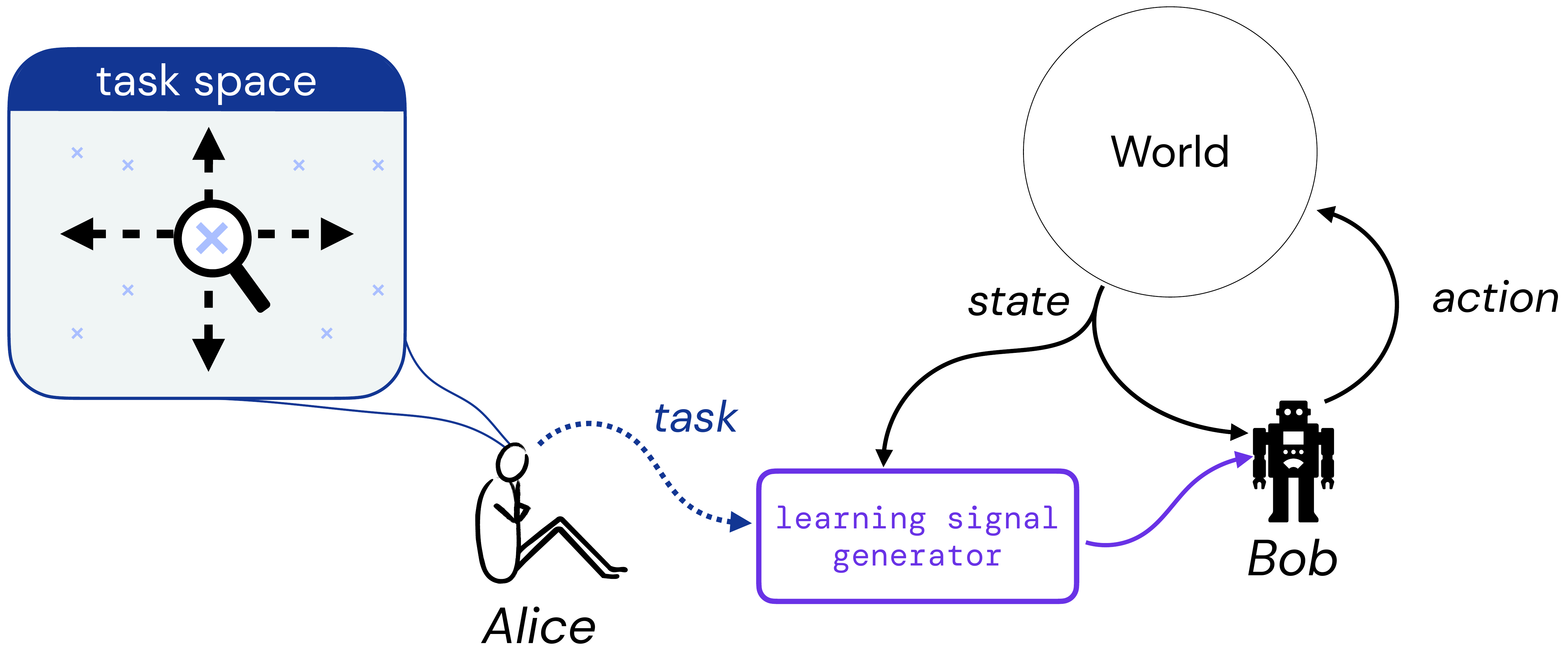}
    \caption{Alice, Bob, and the artifacts of task definition (blue) and task expression (purple).}
    \label{fig:alice_and_bob}
\end{figure}

%
\paragraph{1. An Account of ``Task''.} As rewards encode tasks, goals, or desires, we first ask, ``what \emph{is} a task?''. We frame our study around a thought experiment (\autoref{fig:alice_and_bob}) involving the interactions between a designer, Alice, and a learning agent, Bob, drawing inspiration from \citet{ackley1992interactions}, \citet{sorg2011optimal}, and \citet{singh2009rewards}. In this thought experiment, we draw a distinction between how Alice thinks of a task (\taskq) and the means by which Alice incentivizes Bob to pursue this task (\exprq). This distinction allows us to analyze the expressivity of reward as an answer to the latter question, conditioned on how we answer the former.
%
Concretely, we study three answers to the \taskq\ in the context of finite Markov Decision Processes (MDPs): A task is either (1) a set of acceptable behaviors (policies), (2) a partial ordering over behaviors, or (3) a partial ordering over trajectories. Further detail and motivation for these task types is provided in \autoref{sec:tasq_exprq}, but broadly they can be viewed as generalizations of typical notions of task such as a choice of goal or optimal behavior. Given these three answers to the \taskq, we then examine the \textit{expressivity} of reward.

%

%
\paragraph{2. Expressivity of Markov Reward.} The core of our study asks whether there are tasks Alice would like to convey---as captured by the answers to the \taskq---that admit no characterization in terms of a Markov reward function. Our emphasis on Markov reward functions, as opposed to arbitrary history-based reward functions, is motivated by several factors. First, disciplines such as computer science, psychology, biology, and economics typically rely on a notion of reward as a numerical proxy for the \textit{immediate} worth of states of affairs (such as the financial cost of buying a solar panel or the fitness benefits of a phenotype). Given an appropriate way to describe states of affairs, Markov reward functions can represent immediate worth in an intuitive manner that also allows for reasoning about combinations, sequences, or re-occurrences of such states of affairs. Second, 
it is not clear that general history-based rewards are a reasonable target for learning as they suffer from the curse of dimensionality in the length of the history. 
Lastly, Markov reward functions are the standard in RL. A rigorous analysis of which tasks they can and cannot convey may provide guidance into when it is necessary to draw on alternative formulations of a problem. Given our focus on Markov rewards, we treat a reward function as accurately \textit{expressing} a task just when the value function it induces in an environment adheres to the constraints of a given task.

%
\paragraph{3. Main Results.} We find that, for all three task types, there are environment--task pairs for which there is no Markov reward function that realizes the task (\autoref{thm:main_result_express}). In light of this finding, we design polynomial-time algorithms that can determine, for any given task and environment, whether a reward function exists in the environment that captures the task (\autoref{thm:reward_design_is_in_p}). When such a reward function does exist, the algorithms also return it. Finally, we conduct simple experiments with these procedures to provide empirical insight into the expressivity of reward (\autoref{sec:experiments}).

%
Collectively, our results demonstrate that there are tasks that cannot be expressed by Markov reward in a rigorous sense, but we can efficiently construct such reward functions when they do exist (and determine when they do not). We take these findings to shed light on the nature of reward maximization as a principle, and highlight many pathways for further investigation.

\section{Background}

%
RL defines the problem facing an agent that learns to improve its behavior over time by interacting with its environment. We make the typical assumption that the RL problem is well modeled by an agent interacting with a finite Markov Decision Process (MDP), defined by the tuple $(\mc{S}, \mc{A}, R, T, \gamma, s_0)$. An MDP gives rise to deterministic behavioral policies, $\pi : \mc{S} \ra \mc{A}$, and the value, $V^{\pi} : \mc{S} \ra \mathbb{R}$, and action--value, $Q^{\pi} : \mc{S} \times \mc{A} \ra \mathbb{R}$, functions that measure their quality. We will refer to a Controlled Markov Process (CMP) as an MDP without a reward function, which we denote $E$ for environment. We assume that all reward functions are deterministic, and may be a function of either state, state-action pairs, or state-action-state triples, but \textit{not} history. Henceforth, we simply use ``reward function'' to refer to a deterministic Markov reward function for brevity, but note that more sophisticated settings beyond MDPs and deterministic Markov reward functions are important directions for future work. For more on MDPs or RL, see the books by \citet{puterman2014markov} and \citet{sutton2018reinforcement} respectively.

\subsection{Other Perspectives on Reward}

%
We here briefly summarize relevant literature that provides distinct perspectives on reward.

%
\paragraph{Two Roles of Reward.} 
As \citet{sorg2011optimal} identifies (Chapter 2), reward can both define the task the agent learns to solve, and define the ``bread crumbs'' that allow agents to efficiently learn to solve the task. This distinction has been raised elsewhere \cite{ackley1992interactions,singh2009rewards,singh2010separating}, and is similar to the extrinsic-intrinsic reward divide \cite{singh2005intrinsically,zheng2020can}. Tools such as reward design \cite{mataric1994reward,sorg2010reward} or reward shaping \cite{ng1999policy} focus on offering more efficient learning in a variety of environments, so as to avoid issues of sparsity and long-term credit assignment. We concentrate primarily on reward's capacity to express a \textit{task}, and defer learning dynamics to an (important) stage of future work.

%
\paragraph{Discounts, Expectations, and Rationality.} Another important facet of reward is how it is used in producing behavior. The classical view offered by the Bellman equation (and the reward hypothesis) is that the quantity of interest to maximize is expected, discounted, cumulative reward. Yet it is possible to disentangle reward from the expectation \cite{bellemare2017distributional}, to attend only to ordinal \cite{weng2011markov} or maximal rewards \cite{krishna2020maximum}, or to adopt different forms of discounting \cite{white2017unifying,fedus2019hyperbolic}. In this work, we take the standard view that agents will seek to maximize \textit{value} for a particular discount factor $\gamma$, but recognize that there are interesting directions beyond these commitments, such as inspecting the limits of reward in constrained MDPs as studied by \citet{csabaRLhypothesis}. We also note the particular importance of work by \citet{pitis2019rethinking}, who examines the relationship between classical decision theory \cite{vonneumann1953theory} and MDPs by incorporating additional axioms that account for stochastic processes with discounting \cite{koopmans1960stationary,mitten1974preference,sobel1975ordinal,sobel2013discounting}. Drawing inspiration from \citet{pitis2019rethinking} and \citet{sunehag2011axioms}, we foresee valuable pathways for future work that further makes contact between RL and various axioms of rationality.

%
%

%
\paragraph{Preferences.} In place of numerical rewards, preferences of different kinds may be used to evaluate an agent's behaviors, drawing from the literature on preference-learning \cite{kreps1988notes} and ordinal dynamic programming \cite{debreu1954representation,mitten1974preference,sobel1975ordinal}. This premise gives rise to \textit{preference-based reinforcement learning} (PbRL) in which an agent interacts with a CMP and receives evaluative signals in the form of preferences over states, actions, or trajectories. This kind of feedback inspires and closely parallels the task types we propose in this work. A comprehensive survey of PbRL by \citet{wirth2017survey} identifies critical differences in this setup from traditional RL, categorizes recent algorithmic approaches, and highlights important open questions. Recent work focuses on analysing the sample efficiency of such methods~\cite{xu2020preference,novoseller2020dueling} with close connections to learning from human feedback in real time~\cite{knox2009interactively,macglashan2016convergent,christiano2017deep}.

%
\paragraph{Teaching and Inverse RL.}
The inverse RL (IRL) and apprenticeship learning literature examine the problem of learning directly from behavior \cite{ng2000algorithms,abbeel2004apprenticeship}. The classical problem of IRL is to identify which reward function (often up to an equivalence class) a given demonstrator is optimizing. We emphasize the relevance of two approaches: First, work by \citet{syed2008apprenticeship}, who first illustrate the applicability of linear programming \cite{karmarkar1984new} to apprenticeship learning; and second, work by \citet{amin2017repeated}, who examine the \textit{repeated} form of IRL. The methods of IRL have recently been expanded to include variations of \textit{cooperative} IRL \cite{hadfield2016cooperative}, and \textit{assistive} learning \cite{shah2021benefits}, which offer different perspectives on how to frame interactive learning problems.

%
\paragraph{Reward Misspecification.} 
Reward is also notoriously hard to specify. As pointed out by \citet{littman2017environment}, ``putting a meaningful dollar figure on scuffing a wall or dropping a clean fork is challenging.'' Along these lines, \citet{hadfield2017inverse} identify cases in which well-intentioned designers create reward functions that produce unintended behavior \cite{ortega2018}. \citet{macglashan2017interactive} find that human-provided rewards tend to depend on a learning agent's entire policy, rather than just the current state. Further, work by \citet{hadfield2016off} and \citet{kumar2020realab} suggest that there are problems with reward as a learning mechanism due to misspecification and reward tampering \cite{everitt2017reinforcement}. These problems have given rise to approaches to \textit{reward learning}, in which a reward function is inferred from some evidence such as behavior or comparisons thereof \cite{jeon2020reward}.

%
\paragraph{Other Notions of Task.} As a final note, we highlight alternative approaches to task specification. Building on the Free Energy Principle \cite{friston2009reinforcement,friston2010free}, \citet{hafner2020action} consider a variety of task types in terms of minimization of distance to a desired target distribution \cite{akshay2013steady}. Alternatively, \citet{littman2017environment} and \citet{li2017reinforcement} propose variations of \textit{linear temporal logic} (LTL) as a mechanism for specifying a task to RL agents, with related literature extending LTL to the multi-task \cite{toro2018teaching} and multi-agent \cite{hammond2021multi} settings, or using reward machines for capturing task structure \cite{icarte2018using}. \citet{jothimurugan2020composable} take a similar approach and propose a task specification language for RL based on logical formulas that evaluate whether trajectories satisfy the task, similar in spirit to the logical task compositions framework developed by \citet{tasse2020boolean}. Many of these notions of task are more general than those we consider. A natural direction for future work broadens our analysis to include these kinds of task.

\section{An Account of Reward's Expressivity: The \taskq\ and \exprq}
\label{sec:tasq_exprq}

%
Consider an onlooker, Alice, and an earnest learning agent, Bob, engaged in the interaction pictured in \autoref{fig:alice_and_bob}. Suppose that Alice has a particular task in mind that she would like Bob to learn to solve, and that Alice constructs a reward function to incentivize Bob to pursue this task. Here, Alice is playing the role of ``all of what we mean by goals and purposes'' for Bob to pursue, with Bob playing the role of the standard reward-maximizing RL agent.

%
\paragraph{Two Questions About Task.} To give us leverage to study the expressivity of reward, it is useful to draw a distinction between two stages of this process: 1) Alice thinks of a task that she would like Bob to learn to solve, and 2) Alice creates a reward function (and perhaps chooses $\gamma$) that conveys the chosen task to Bob. We inspect these two separately, framed by the following two questions. The first we call the \textit{task-definition question} (\taskq) which asks: What \textit{is} a task? The second we call the \textit{task-expression question} (\exprq) which asks: Which learning signal can be used as a mechanism for expressing any task to Bob?

%
\paragraph{Reward Answers The \exprq.} We suggest that it may be useful to treat reward as an answer to the \exprq\ rather than the \taskq. 
On this view, reward is treated as an expressive language for incentivizing reward-maximizing agents: Alice may attempt to translate any task into a reward function that incentivizes Bob to pursue the task, no matter which environment Bob inhabits, which task Alice has chosen, or how she has represented the task to herself.
Indeed, it might be the case that Alice's knowledge of the task far exceeds Bob's representational or perceptual capacity. Alice may know every detail of the environment and define the task based on this holistic vantage, while Bob must learn to solve the task through interaction alone, relying only on a restricted class of functions for modeling and decision making. 

Under this view, we can assess the expressivity of reward as an answer to the \exprq\ conditioned on how we answer the \taskq.
%
%
For example, if the \taskq\ is answered in terms of natural language descriptions of desired states of affairs, then reward may fail to convey the chosen task due to the apparent mismatch in abstraction between natural language and reward (though some work has studied such a proposal \cite{macglashan15,williams2018learning}).

\subsection{Answers to the \taskq: What is a Task?}

%
In RL, tasks are often associated with a choice of goal, reward function ($R$), reward-discount pair ($R,\gamma$), or perhaps a choice of optimal policy (alongside those task types surveyed previously, such as LTL). However, it is unclear whether these constructs capture the entirety of what we mean by ``task''.

%
For example, consider the \citet{russell94} grid world: A 4$\times$3 grid with one wall, one terminal fire state, and one terminal goal state (pictured with a particular reward function in \autoref{subfig:soap_rew_visual}). In such an environment, how might we think about tasks? A standard view is that the task is to reach the goal as quickly as possible. This account, however, fails to distinguish between the {\em non-}optimal behaviors, such as the costly behavior of the agent moving directly into the fire and the neutral behavior of the agent spending its existence in the start state. Indeed, characterizing a task in terms of choice of $\pi^*$ or goal fails to capture these distinctions. 
Our view is that a suitably rich account of task should allow for the characterization of this sort of preference, offering the flexibility to scale from specifying only the desirable behavior (or outcomes) to an arbitrary ordering over behaviors (or outcomes).

%
In light of these considerations, we propose three answers to the \taskq\ that can convey general preferences over behavior or outcome: 1) A set of acceptable policies, 2) A partial ordering over policies, or 3) A partial ordering over trajectories. We adopt these three as they can capture many kinds of task while also allowing a great deal of flexibility in the level of detail of the specification.

\subsection{SOAPs, POs, and TOs}

%
\paragraph{(SOAP) Set Of Acceptable Policies.} A classical view of the equivalence of two reward functions is based on the optimal policies they induce. For instance, \citet{ng1999policy} develop potential-based reward shaping by inspecting which shaped reward signals will ensure that the optimal policy is unchanged. Extrapolating, it is natural to say that for any environment $E$, two reward functions are equivalent if the optimal policies they induce in $E$ are the same. 
In this way, a task is viewed as a choice of optimal policy. As discussed in the grid world example above, this notion of task fails to allow for the specification of the quality of other behaviors. For this reason, we generalize task-as-optimal-policy to a \textit{set of acceptable policies}, defined as follows.

%
\begin{definition}
A set of acceptable policies (SOAP) is a non-empty subset of the deterministic policies, $\Pi_G \subseteq \Pi$, with $\Pi$ the set of all deterministic mappings from $\mc{S}$ to $\mc{A}$ for a given $E$.
\end{definition}

%
With one task type defined, it is important to address what it means for a reward function to properly \textit{realize}, \textit{express}, or \textit{capture} a task in a given environment. We offer the following account.

%
\begin{definition}
A reward function is said to \textit{realize} a task $\mathscr{T}$ in an environment $E$ just when the start-state value (or return) induced by the reward function exactly adheres to the constraints of $\mathscr{T}$.
\end{definition}
Precise conditions for the realization of each task type are provided alongside each task definition, with a summary presented in column four of \autoref{tab:task_types}.

%
For SOAPs, we take the start-state value $V^{\pi}(s_0)$ to be the mechanism by which a reward function realizes a SOAP. That is, for a given $E$ and $\Pi_G$, a reward function $R$ is said to \textit{realize} the $\Pi_G$ in $E$ when the start-state value function is optimal for all good policies, and strictly higher than the start-state value of all other policies. It is clear that SOAP strictly generalizes a task in terms of a choice of optimal policy, as captured by the SOAP $\Pi_G = \{\pi^*\}$.

%
We note that there are two natural ways for a reward function to realize a SOAP: First, each $\pi_g \in \Pi_G$ has \textit{optimal} start-state value and all other policies are sub-optimal. We call this type \textit{equal}-SOAP, or just SOAP for brevity. Alternatively, we might only require that the acceptable policies are each \textit{near}-optimal, but are allowed to differ in start-state value so long as they are all better than \textit{every} bad policy $\pi_b \in \Pi_B$. That is, in this second kind, there exists an $\epsilon \geq 0$ such that every $\pi_g \in \Pi_G$ is $\epsilon$-optimal in start-state value, $V^*(s_0) - V^{\pi_g}(s_0) \leq \epsilon,$ while all other policies are worse. We call this second realization condition \textit{range}-SOAP. We note that the range realization generalizes the equal one: Every equal-SOAP is a range-SOAP (by letting $\epsilon=0$). However, there exist range-SOAPs that are expressible by Markov rewards that are \textit{not} realizable as an equal-SOAP. We illustrate this fact with the following proposition. All proofs are presented in \autoref{apend:proofs}. 

%
\begin{proposition}
\label{prop:soap_separation}
There exists a CMP, $E$, and choice of $\Pi_G$ such that $\Pi_G$ can be realized under the range-SOAP criterion, but cannot be realized under the equal-SOAP criterion.
\end{proposition}

%
One such CMP is pictured \autoref{subfig:entail}. Consider the SOAP $\Pi_G = \{\pi_{11}, \pi_{12}, \pi_{21}\}$: Under the equal-SOAP criterion, if each of these three policies are made optimal, any reward function will \textit{also} make $\pi_{22}$ (the only bad policy) optimal as well. In contrast, for the range criterion, we can choose a reward function that assigns lower rewards to $a_2$ than $a_1$ in both states. 
In general, we take the equal-SOAP realization as canonical, as it is naturally subsumed by our next task type.

%
\input{tables/task_types}

%
\paragraph{(PO) Partial Ordering on Policies.} Next, we suppose that Alice chooses a \textit{partial ordering} on the deterministic policy space. That is, Alice might identify a some great policies, some good, and some bad policies to strictly avoid, and remain indifferent to the rest. POs strictly generalize equal SOAPs, as any such SOAP is a special choice of PO with only two equivalence classes. We offer the following definition of a PO.

%
\begin{definition}
A policy order (PO) of the deterministic policies $\Pi$ is a partial order, denoted $L_{\Pi}$.
\end{definition}

%
As with SOAPs, we take the start-state value $V^{\pi}(s_0)$ induced by a reward function $R$ as the mechanism by which policies are ordered. That is, given $E$ and $L_\Pi$, we say that a reward function $R$ \textit{realizes} $L_\Pi$ in $E$ if and only if the resulting MDP, $M = (E, R)$, produces a start-state value function that orders $\Pi$ according to $L_\Pi$.

%
\paragraph{(TO) Partial Ordering on Trajectories.} A natural generalization of goal specification enriches a notion of task to include the details of how a goal is satisfied---that is, for Alice to relay some preference over trajectory space \cite{wilson2012bayesian}, as is done in preference based RL \cite{wirth2017survey}. Concretely, we suppose Alice specifies a partial ordering on length $N$ trajectories of $(s,a)$ pairs, defined as follows.

%
\begin{definition}
A trajectory ordering (TO) of length $N \in \N$ is a partial ordering $L_{\tau,N}$, with each trajectory $\tau$ consisting of $N$ state--action pairs, $\{(s_0, a_0), \ldots, (a_{N-1}, s_{N-1})\}$, with $s_0$ the start state.
\end{definition}

%
As with PO, we say that a reward function realizes a trajectory ordering $L_{\tau, N}$ if the ordering determined by each trajectory's cumulative discounted $N$-step return from $s_0$, denoted $G(\tau; s_0)$, matches that of the given $L_{\tau, N}$. We note that trajectory orderings can generalize goal-based tasks at the expense of a larger specification. For instance, a TO can convey the task, ``Safely reach the goal in less than thirty steps, or just get to the subgoal in less than twenty steps.''

%
\paragraph{Recap.} We propose to assess the expressivity of reward by first answering the \taskq\ in terms of SOAPs, POs, or TOs, as summarized by \autoref{tab:task_types}. We say that a task $\mathscr{T}$ is \textit{realized} in an environment $E$ under reward function $R$ if the start-state value function (or return) produced by $R$ imposes the constraints specified by $\mathscr{T}$, and are interested in whether reward can always realize a given task in any choice of $E$. 
%
We make a number of assumptions along the way, including: (1) Reward functions are Markov and deterministic, (2) Policies of interest are deterministic, 
(3) The environment is a finite CMP, (4) $\gamma$ is part of the environment, (5) We ignore reward's role in shaping the \textit{learning process}, (6) Start-state value or return is the appropriate mechanism to determine if a reward function realizes a given task. Relaxation of these assumptions is a critical direction for future work.

\section{Analysis: The Expressivity of Markov Reward}
\label{sec:expressivity}

%
With our definitions and objectives in place, we now present our main results.

\subsection{Express SOAPs, POs, and TOs}

We first ask whether reward can always realize a given SOAP, PO, or TO, for an arbitrary $E$. Our first result states that the answer is ``no''---there are tasks that cannot be realized by any reward function. 

\begin{theorem}
For each of SOAP, PO, and TO, there exist $(E,\mathscr{T})$ pairs for which no Markow reward function realizes $\mathscr{T}$ in $E$.
\label{thm:main_result_express}
\end{theorem}

%
Thus, reward is incapable of capturing certain tasks. What tasks are they, precisely? 
Intuitively, inexpressible tasks involve policies or trajectories that \textit{must} be correlated in \textit{value} in an MDP. That is, if two policies are nearly identical in behavior, it is unlikely that reward can capture the PO that places them at opposite ends of the ordering. A simple example is the ``always move the same direction'' task in a grid world, with state defined as an $(x,y)$ pair. The SOAP $\Pi_G = \{\pi_{\la}, \pi_{\uparrow}, \pi_{\ra}, \pi_{\downarrow}\}$ conveys this task, but no Markov reward function can make these policies strictly higher in value than all others. 

%
\begin{figure}[!t]
    \centering
    \subfloat[Steady State Case]{\includegraphics[width=0.4\textwidth]{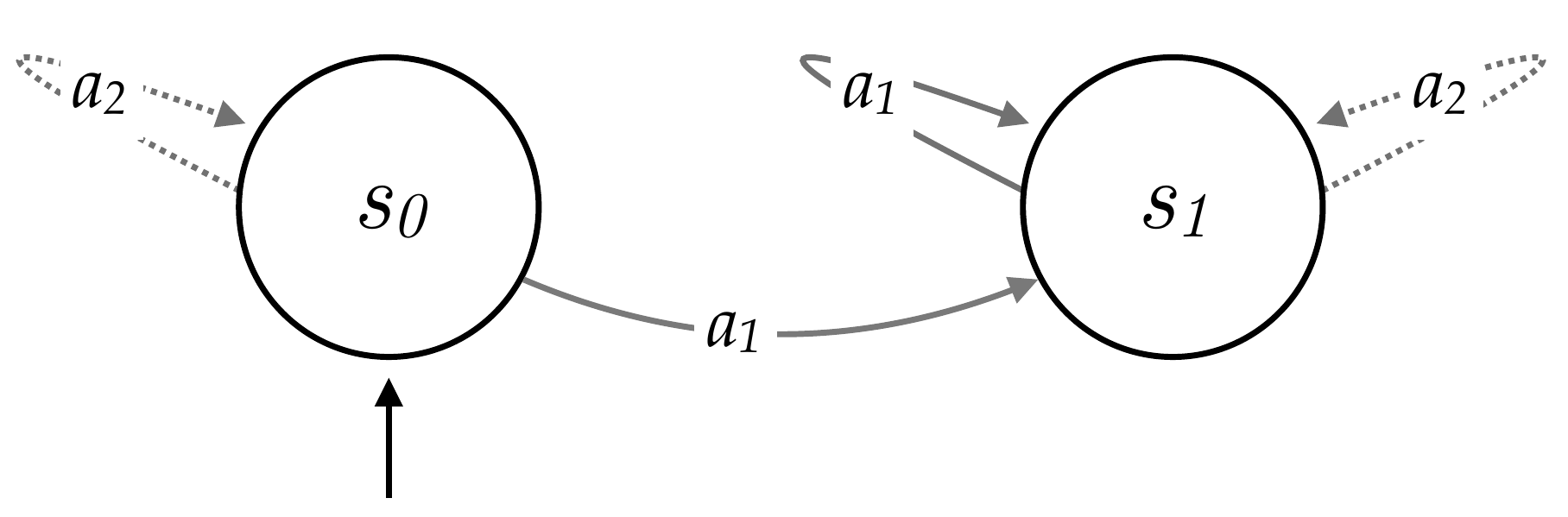}} \hspace{5mm}
    \subfloat[Entailment Case\label{subfig:entail}]{\includegraphics[width=0.4\textwidth]{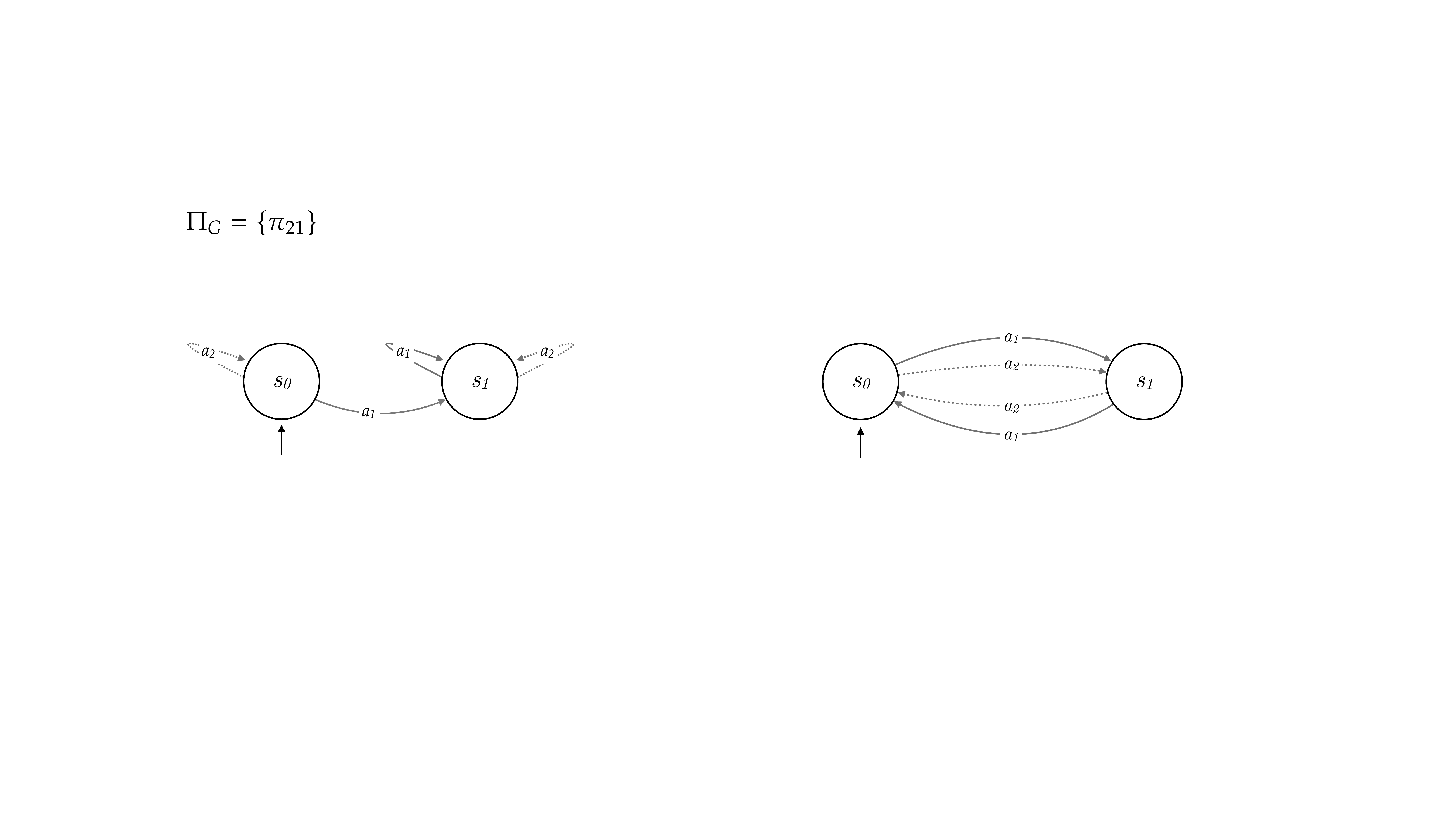}}
    \caption{Two CMPs in which there is a SOAP that is not expressible under any Markov reward function. On the left, $\Pi_G = \{\pi_{21}\}$ is not realizable, as $\pi_{21}$ can not be made better than $\pi_{22}$ because $s_1$ is never reached. On the right, the XOR-like-SOAP, $\Pi_G = \{\pi_{12}, \pi_{21}\}$ is not realizable: To make these two policies optimal, it is entailed that $\pi_{22}$ and $\pi_{11}$ must be optimal, too.}
    \label{fig:example_inexpr_soap}
\end{figure}

%
\paragraph{Example: Inexpressible SOAPs.} Observe the two CMPs pictured in \autoref{fig:example_inexpr_soap}, depicting two kinds of inexpressible SOAPs. On the left, we consider the SOAP $\Pi_G = \{\pi_{21}\}$, containing only the policy that executes $a_2$ in the left state ($s_0$), and $a_1$ in the right ($s_1$). This SOAP is inexpressible through reward, but only because reward cannot distinguish the start-state value of $\pi_{21}$ and $\pi_{22}$ since the policies differ only in an unreachable state. This is reminiscent of Axiom 5 from \citet{pitis2019rethinking}, which explicitly excludes preferences of this sort. On the right, we find a more interesting case: The chosen SOAP is similar to the XOR function, $\Pi_G = \{\pi_{12},\pi_{21}\}$. Here, the task requires that the agent choose each action in exactly one state. However, there cannot exist a reward function that makes \textit{only} these policies optimal, as by consequence, both policies $\pi_{11}$ and $\pi_{22}$ \textit{must} be optimal as well. 

%
Next, we show that \autoref{thm:main_result_express} is not limited to a particular choice of transition function or $\gamma$.
\begin{proposition}
There exist choices of $E_{\neg T} = (\mc{S}, \mc{A}, \gamma, s_0)$ or $E_{\neg \gamma} = (\mc{S}, \mc{A}, T, s_0)$, together with a task $\mathscr{T}$, such that there is no $(T,R)$ pair that realizes $\mathscr{T}$ in $E_{\neg T}$ or $(R,\gamma)$ in $E_{\neg \gamma}$.
\label{prop:fixed_sag_vary_t_r}
\end{proposition}
This result suggests that the scope of \autoref{thm:main_result_express} is actually quite broad---even if the transition function or $\gamma$ are taken as part of the reward specification, there are tasks that cannot be expressed. We suspect there are ways to give a precise characterization of \textit{all} inexpressible tasks from an axiomatic perspective, which we hope to study in future work.

\subsection{Constructive Algorithms: Task to Reward}

%
We now analyze how to determine whether an appropriate reward function can be constructed for any $(E,\mathscr{T})$ pair. We pose a general form of the reward-design problem \cite{mataric1994reward,sorg2010reward,dewey2014reinforcement} as follows.

%
\begin{definition}
The \textsc{RewardDesign} problem is: \textbf{Given} $E = (\mc{S}, \mc{A}, T, \gamma, s_0)$, and a $\mathscr{T}$, \textbf{output} a reward function $R_{\text{alice}}$ that ensures $\mathscr{T}$ is realized in $M = (E, R_{\text{alice}})$.
\end{definition}

%
Indeed, for all three task types, there is an efficient algorithm for solving the reward-design problem.

\begin{theorem}
The \textsc{RewardDesign} problem can be solved in polynomial time, for any finite $E$, and any $\mathscr{T}$, so long as reward functions with infinitely many outputs are considered.
\label{thm:reward_design_is_in_p}
\end{theorem}
%

%
Therefore, for \textit{any} choice of finite CMP, $E$, and a SOAP, PO, or TO, we can find a reward function that perfectly realizes the task in the given environment, if such a reward function exists. Each of the three algorithms are based on forming a linear program that matches the constraints of the given task type, which is why reward functions with infinitely many outputs are required. Pseudo-code for SOAP-based reward design is presented in \autoref{alg:soap_rew_design}. Intuitively, the algorithms compute the discounted expected-state visitation distribution for a collection of policies; in the case of SOAP, for instance, these policies include $\Pi_G$ and what we call the ``fringe'', the set of policies that differ from a $\pi_g \in \Pi_G$ by exactly one action. Then, we use these distributions to describe linear inequality constraints ensuring that the start-state value of the good policies are better than those of the fringe.

%
\input{algorithms/soap_reward_design}


%
As highlighted by \autoref{thm:main_result_express} there are SOAPs, POs, and TOs that are \textit{not realizable}. Thus, it is important to determine how the algorithms mentioned in \autoref{thm:reward_design_is_in_p} will handle such cases. Our next corollary illustrates that the desirable outcome is achieved: For any $E$ and $\mathscr{T}$, the algorithms will output a reward function that realizes $\mathscr{T}$ in $E$, or output `$\perp$' when no such function exists.

\begin{corollary}
For any task $\mathscr{T}$ and environment $E$, deciding whether $\mathscr{T}$ is expressible in $E$ is solvable in polynomial time.
\label{cor:rd_decidable}
\end{corollary}

%
Together, \autoref{thm:main_result_express} and \autoref{thm:reward_design_is_in_p} constitute our main results: There are environment--task pairs in which Markov reward cannot express the chosen task for each of SOAPs, POs, and TOs. However, there are efficient algorithms for deciding whether a task is expressible, and for constructing the realizing reward function when it exists. We will study the use of one of these algorithms in \autoref{sec:experiments}, but first attend to other aspects of reward's expressivity.

\subsection{Other Aspects of Reward's Expressivity}

%
We next briefly summarize other considerations about the expressivity of reward. 
As noted, \autoref{thm:reward_design_is_in_p} requires the use of a reward function that can produce infinitely many outputs. Our next result proves this requirement is strict for efficient reward design.

%
\begin{theorem}
A variant of the \textsc{RewardDesign} problem with finite reward outputs is NP-hard.
\label{thm:finite_po_np-hard}
\end{theorem}

%
We provide further details about the precise problem studied in \autoref{apend:proofs}. 
Beyond reward functions with finitely-many outputs, we are also interested in extensions of our results to multiple \textit{environments}. We next present a positive result indicating our algorithms can extend to the case where Alice would like to design a reward function for a single task across multiple environments.

%
\begin{proposition}
For any SOAP, PO, or TO, given a finite set of CMPs, $\mc{E} = \{E_1, \ldots, E_n\}$, with shared state--action space, there exists a polynomial time algorithm that outputs one reward function that realizes the task (when possible) in all CMPs in $\mc{E}$.
\label{prop:multi_environment}
\end{proposition}

%
A natural follow up question to the above result asks whether task realization is \textit{closed} under a set of CMPs. Our next result answers this question in the negative. 
%
\begin{theorem}
Task realization is not closed under sets of CMPs with shared state-action space. That is, there exist choices of $\mathscr{T}$ and $\mc{E} = \{E_1, \ldots, E_n\}$ such that $\mathscr{T}$ is realizable in each $E_i \in \mc{E}$ independently, but there is not a single reward function that realizes $\mathscr{T}$ in all $E_i \in \mc{E}$ simultaneously.
\label{thm:realize_not_closed}
\end{theorem}
Intuitively, this shows that Alice must know precisely which environment Bob will inhabit if she is to design an appropriate reward function. Otherwise, her uncertainty over $E$ may prevent her from designing a realizing reward function. We foresee iterative extensions of our algorithms in which Alice and Bob can react to one another, drawing inspiration from repeated IRL by \citet{amin2017repeated}.

\section{Experiments}
\label{sec:experiments}

%
We next conduct experiments to shed further light on the findings of our analysis. Our focus is on SOAPs, though we anticipate the insights extend to POs and TOs as well with little complication. In the first experiment, we study the fraction of SOAPs that are expressible in small CMPs as we vary aspects of the environment or task (\autoref{fig:results_soap_expressivity}). In the second, we use one algorithm from \autoref{thm:reward_design_is_in_p} to design a reward function, and contrast learning curves under a SOAP-designed reward function compared to standard rewards. Full details about the experiments are found in \autoref{apend:experiments}. 

\input{figures/results_1_expressivity}

%
\paragraph{SOAP Expressivity.} First, we estimate the fraction of SOAPs that are expressible in small environments. For each data point, we sample 200 random SOAPs and run \autoref{alg:soap_rew_design} described by \autoref{thm:reward_design_is_in_p} to determine whether each SOAP is realizable in the given CMP. We ask this question for both the equal (color) variant of SOAP realization and the range (grey) variant. We inspect SOAP expressivity as we vary six different characteristics of $E$ or $\Pi_G$: The number of actions, the number of states, the discount $\gamma$, the number of good policies in each SOAP, the Shannon entropy of $T$ at each $(s,a)$ pair, and the ``spread'' of each SOAP. The spread approximates average edit distance among policies in $\Pi_G$ determined by randomly permuting actions of a reference policy by a coin weighted according to the value on the x-axis. We use the same set of CMPs for each environment up to any deviations explicitly made by the varied parameter (such as $\gamma$ or entropy). Unless otherwise stated, each CMP has four states and three actions, with a fixed but randomly chosen transition function.

%
Results are presented in \autoref{fig:results_soap_expressivity}. We find that our theory is borne out in a number of ways. First, as \autoref{thm:main_result_express} suggests, we find SOAP expressivity is strictly less than one in nearly all cases. This is evidence that inexpressible tasks are not only found in manufactured corner cases, but rather that expressivity is a spectrum. We further observe---as predicted by \cref{prop:soap_separation}---clear separation between the expressivity of range-SOAP (grey) vs. equal-SOAP (color); there are many cases where we can find a reward function that makes the good policies \textit{near} optimal and better than the bad, but cannot make those good policies all \textit{exactly} optimal. Additionally, several trends emerge as we vary the parameter of environment or task, though we note that such trends are likely specific to the choice of CMP and may not hold in general. Perhaps the most striking trend is in \autoref{subfig:spread}, which shows a decrease in expressivity as the SOAPs become more \textit{spread out}. This is quite sensible: A more spread out SOAP is likely to lead to more entailments of the kind discussed in \autoref{subfig:entail}.

%
\paragraph{Learning with SOAP-designed Rewards.} Next, we contrast the learning performance of Q-learning under a SOAP-designed reward function (visualized in \autoref{subfig:soap_rew_visual}) with that of the regular goal-based reward in the \citet{russell94} grid world. In this domain, there is 0.35 slip probability such that, on a `slip' event, the agent randomly applies one of the two orthogonal action effects.
The regular goal-based reward function provides $+1$ when the agent enters the terminal flag cell, and $-1$ when the agent enters the terminal fire cell. The bottom left state is the start-state, and the black cell is an impassable wall.

%
Results are presented in \autoref{fig:rn_grid_results}. On the right, we present a particular kind of learning curve contrasting the performance of Q-learning with the SOAP reward (blue) and regular reward (green). The y-axis measures, at the end of each episode, the average (inverse) minimum edit distance between Q-learning's greedy policy and any policy in the SOAP. Thus, when the series reaches 1.0, Q-learning's greedy policy is identical to one of the two SOAP policies. We first find that Q-learning is able to quickly learn a $\pi_g \in \Pi_G$ under the designed reward function. We further observe that the typical reward does not induce a perfect match in policy---at convergence, the green curve hovers slightly below the blue, indicating that the default reward function is incentivizing different policies to be optimal. This is entirely sensible, as the two SOAP policies are extremely cautious around the fire; they choose the orthogonal (and thus, safe) action in fire-adjacent states, relying on slip probability to progress. Lastly, as expected given the amount of knowledge contained in the SOAP, the SOAP reward function allows Q-learning to rapidly identify a good policy compared to the typical reward.

\input{figures/results_2_learning}

\section{Conclusion}
\label{sec:conclusion}

%
We have here investigated the expressivity of Markov reward, framed around three new accounts of task. Our main results show that there exist choices of task and environment in which Markov reward cannot express the chosen task, but there are efficient algorithms that decide whether a task is expressible \textit{and} construct a reward function that captures the task when such a function exists. We conclude with an empirical examination of our analysis, corroborating the findings of our theory. We take these to be first steps toward understanding the full scope of the reward hypothesis.

%
There are many routes forward. A key direction moves beyond the task types we study here, and relaxes our core assumptions---the environment might not be a finite CMP, Alice may not know the environment precisely, reward may be a function of history, or Alice may not know how Bob represents state. Along similar lines, a critical direction incorporates how reward impacts Bob's \textit{learning dynamics} rather than start-state value. Further, we note the potential relevance to the recent \textit{reward-is-enough} hypothesis proposed by \citet{silver2021reward}; we foresee pathways to extend our analysis to examine this newer hypothesis, too. For instance, in future work, it is important to assess whether reward is capable of inducing the right kinds of attributes of cognition, not just behavior. 

\begin{ack}
The authors would like to thank Andr{\' e} Barreto, Diana Borsa, Michael Bowling, Wilka Carvalho, Brian Christian, Jess Hamrick, Steven Hansen, Zac Kenton, Ramana Kumar, Katrina McKinney, R{\'e}mi Munos, Matt Overlan, Hado van Hasselt, and Ben Van Roy for helpful discussions. We would also like to thank the anonymous reviewers for their thoughtful feedback, and Brendan O'Donoghue for catching a typo in the appendix. Michael Littman was supported in part by funding from DARPA L2M, ONR MURI, NSF FMitF, and NSF RI.
\end{ack}

\bibliographystyle{plainnat}
\bibliography{reward_hyp}

\input{neurips_appendix-content}

\end{document}

%% file: styles/neurips_commands.tex
\definecolor{dblue}{RGB}{18, 54, 147}
\definecolor{dmblue}{RGB}{169, 193, 219}
\definecolor{dlblue}{RGB}{216, 235, 255}
\definecolor{dred}{RGB}{195, 112, 113}
\definecolor{partred}{RGB}{172, 60, 56}
\definecolor{dorange}{RGB}{230, 169, 132}
\definecolor{ddgreen}{RGB}{54, 95, 49}
\definecolor{dgreen}{RGB}{118, 167, 125}
\definecolor{dlgreen}{RGB}{154, 195, 157}
\definecolor{dtan}{RGB}{221, 215, 200}
\definecolor{dpink}{RGB}{207, 166, 208}
\definecolor{dyellow}{RGB}{255, 248, 199}
\definecolor{dgray}{RGB}{46, 49, 49}

\hypersetup{
    colorlinks=true,
    urlcolor=dblue,
    linkcolor=ddgreen,
    citecolor=dblue
}

\newcommand{\mc}[1]{\mathcal{#1}}
\newcommand{\eps}{\varepsilon}

\newcommand{\ra}{\rightarrow}
\newcommand{\la}{\leftarrow}
\newcommand{\mb}[1]{\mathbb{#1}}
\newcommand{\N}{\mb{N}}

\newtheorem{theorem}{Theorem}
\newtheorem{corollary}[theorem]{Corollary}

\newtheorem{proposition}[theorem]{Proposition}
\newtheorem{definition}{Definition}

\numberwithin{equation}{section}
\numberwithin{theorem}{section}
\numberwithin{definition}{section}

\newmdenv[
  topline=false,
  bottomline=false,
  rightline = false,
  leftmargin=8pt,
  rightmargin=0pt,
  skipabove=16pt, 
  innertopmargin=0pt,
  innerbottommargin=0pt
]{innerproof}

\newenvironment{dproof}[1][Proof]{\begin{proof}[\textbf{\textit{Proof of #1.}}] \text{\vspace{2mm}} \begin{innerproof}}{\end{innerproof}\end{proof}\vspace{4mm}}

\usepackage{etoolbox}
\makeatletter
\patchcmd{\NAT@test}{\else \NAT@nm}{\else \NAT@hyper@{\NAT@nm}}{}{}
\makeatother

\newtheorem{innercustomthm}{Theorem}
\newenvironment{customthm}[1]
  {\renewcommand\theinnercustomthm{#1}\innercustomthm}
  {\endinnercustomthm}

\newtheorem{innercustomcor}{Corollary}
\newenvironment{customcor}[1]
  {\renewcommand\theinnercustomcor{#1}\innercustomcor}
  {\endinnercustomcor}
\newtheorem{innercustomprop}{Proposition}
\newenvironment{customprop}[1]
  {\renewcommand\theinnercustomprop{#1}\innercustomprop}
  {\endinnercustomprop}
\newtheorem{innercustomlem}{Lemma}
\newenvironment{customlem}[1]
  {\renewcommand\theinnercustomlem{#1}\innercustomlem}
  {\endinnercustomlem}
  
\newcommand{\spacerule}{\begin{center}\hdashrule{2cm}{1pt}{1pt}\end{center}}

%% file: tables/task_types.tex
\begin{table}[t!]
\def\arraystretch{1.3}
    \centering
    \begin{tabular}{llll}
        \toprule
        %
         \textit{Name}&\textit{Notation }&\textit{Generalizes}&\textit{Constraints Induced by $\mathscr{T}$} \\
        \midrule
        %
        \multirow{ 2}{*}{SOAP} & \multirow{ 2}{*}{$\Pi_G$}& \multirow{ 2}{*}{task-as-$\pi^*$}&equal:  $V^{\pi_g}(s_0) = V^{\pi_{g'}}(s_0) > V^{\pi_b}(s_0), \forall_{\pi_g, \pi_{g'} \in \Pi_G, \pi_b \in \Pi_B}$ \\
        &&&range:  $V^{\pi_g}(s_0) > V^{\pi_b}(s_0), \forall_{\pi_g \in \Pi_G, \pi_b \in \Pi_B}$ \vspace{6pt}\\
        %
        %
        PO& $L_\Pi$&SOAP& $(\pi_1 \oplus \pi_2) \in L_\Pi \implies V^{\pi_1}(s_0) \oplus V^{\pi_2}(s_0)$ \vspace{6pt}\\
        %
        TO& $L_{\tau, N}$& task-as-goal& $(\tau_1 \oplus \tau_2) \in L_{\tau, N} \implies G(\tau_1;s_0) \oplus G(\tau_2;s_0)$ \\
        \midrule
         
    \end{tabular}
    \caption{A summary of the three proposed task types. We further list the constraints that determine whether a reward function \textit{realizes} each task type in an MDP, where we take $\oplus$ to be one of `$<$', `$>$', or `$=$', and $G$ is the discounted return of the trajectory.}
    \label{tab:task_types}
\end{table}

%% file: algorithms/soap_reward_design.tex
\algtext*{EndWhile}
\algtext*{EndIf}
\algtext*{EndFor}
\begin{algorithm}[!t]
\caption{SOAP Reward Design}
\label{alg:soap_rew_design}
\textsc{Input:} $E = (\mc{S}, \mc{A}, T, \gamma, s_0)$, $\Pi_G$. \\
\textsc{Output:} $R$, or $\perp$. \\

\begin{algorithmic}[1]
\State $\Pi_{\text{fringe}} = \texttt{compute\_fringe}(\Pi_G)$
%
\For{$\pi_{g,i} \in \Pi_G$}  \Comment{Compute state-visitation distributions.}
    \State $\rho_{g,i} = \texttt{compute\_exp\_visit}(\pi_{g,i}, E)$
\EndFor
\vspace{2mm}
\For{$\pi_{f,i} \in \Pi_\texttt{fringe}$}
    \State $\rho_{f,i} = \texttt{compute\_exp\_visit}(\pi_{f,i}, E)$
\EndFor
\vspace{2mm}
%
\State $C_{\text{eq}} = \{\}$ \Comment{Make Equality Constraints.}
\For{$\pi_{g,i} \in \Pi_{G}$}
    \State $C_{\text{eq}}.\texttt{add}(\rho_{g,0}(s_0)\cdot X  = \rho_{g,i}(s_0)\cdot X )$
\EndFor
\vspace{2mm}
%
\State $C_{\text{ineq}} = \{\}$ \Comment{Make Inequality Constraints.}
\For{$\pi_{f,j} \in \Pi_{\text{fringe}}$}
    \State $C_{\text{ineq}}.\texttt{add}(\rho_{f,j}(s_0) \cdot X + \epsilon \leq \rho_{g,0}(s_0) \cdot X)$
\EndFor
\vspace{2mm}
%
\State $R_{\text{out}}, \epsilon_{\text{out}} = \texttt{linear\_programming}(\text{obj.}=\max \epsilon, \text{constraints}=C_{\text{ineq}}, C_{\text{eq}})$ \Comment{Solve LP.}
\vspace{4mm}
%
\If{$\epsilon_{\text{out}} > 0$} \Comment{Check if successful.}

\Return $R_{\text{out}}$
\Else\

\Return $\perp$
\EndIf
\end{algorithmic}
\end{algorithm}

%% file: figures/results_1_expressivity.tex
\begin{figure*}[t!]
    \centering
    %
    \subfloat[Vary Num. Actions\label{subfig:res_num_actions}]{\includegraphics[width=0.3\textwidth]{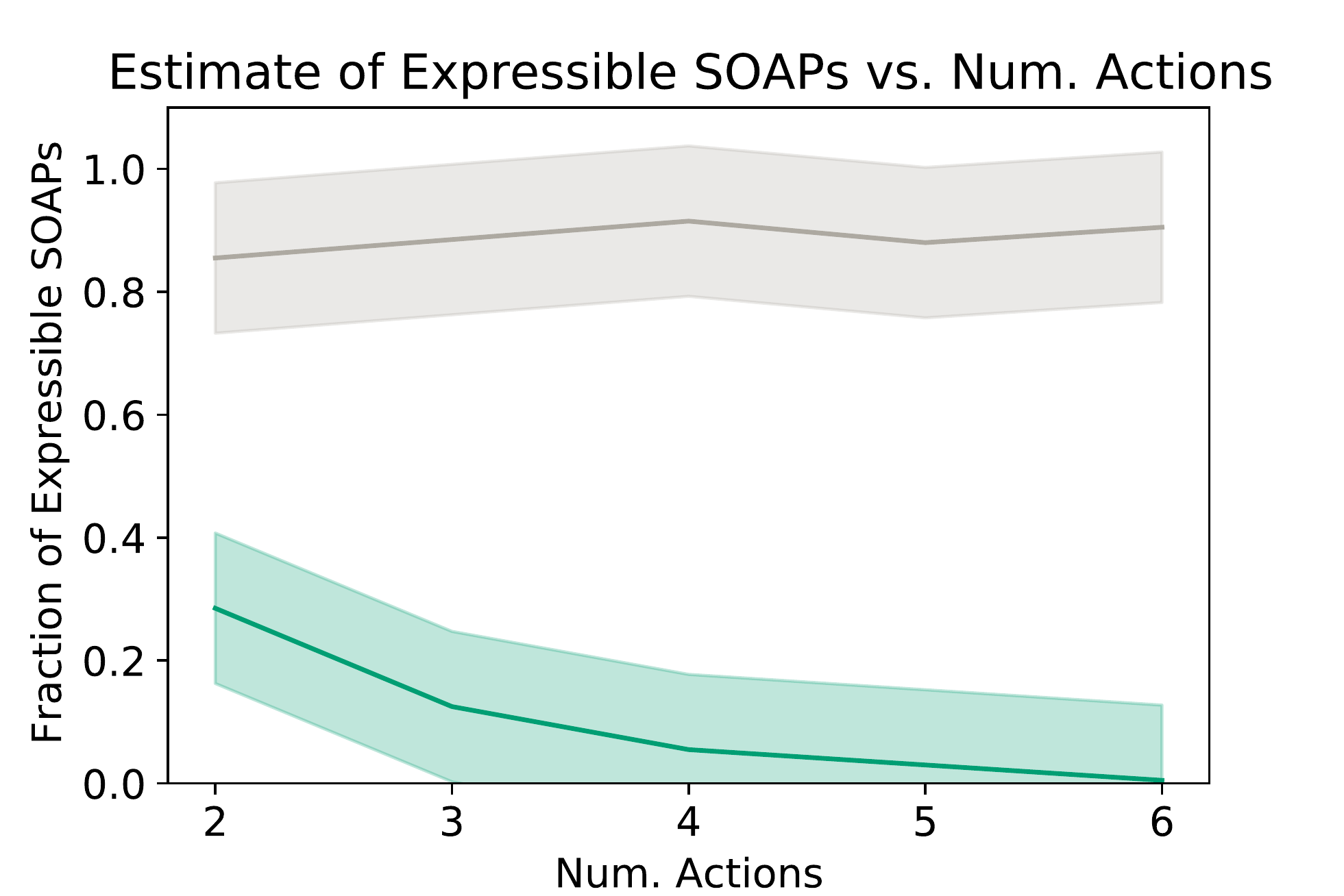}}
    %
    \subfloat[Vary Num. States\label{subfig:res_num_states}]{\includegraphics[width=0.3\textwidth]{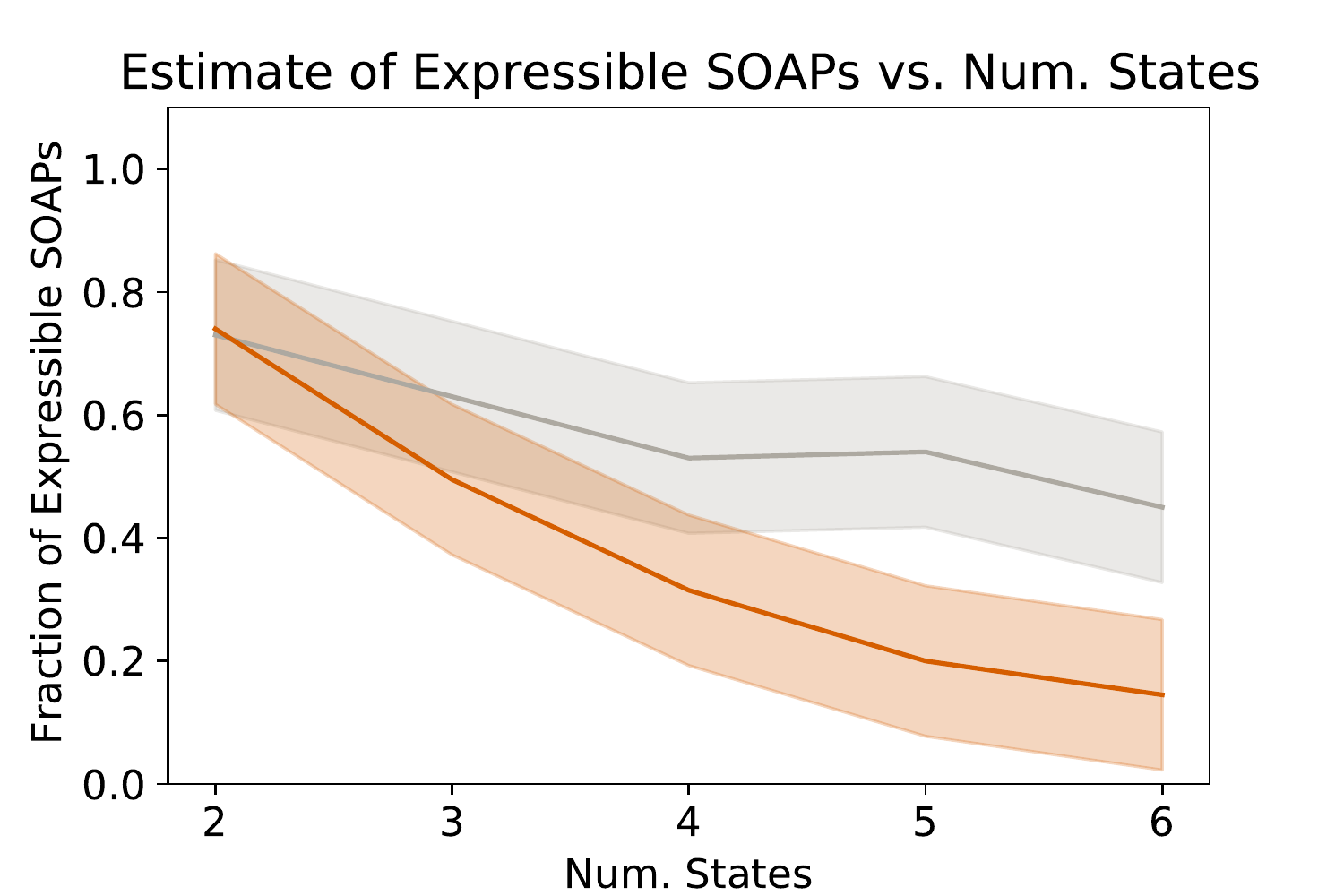}}
    %
    \subfloat[Vary $\gamma$\label{subfig:res_gamma}]{\includegraphics[width=0.3\textwidth]{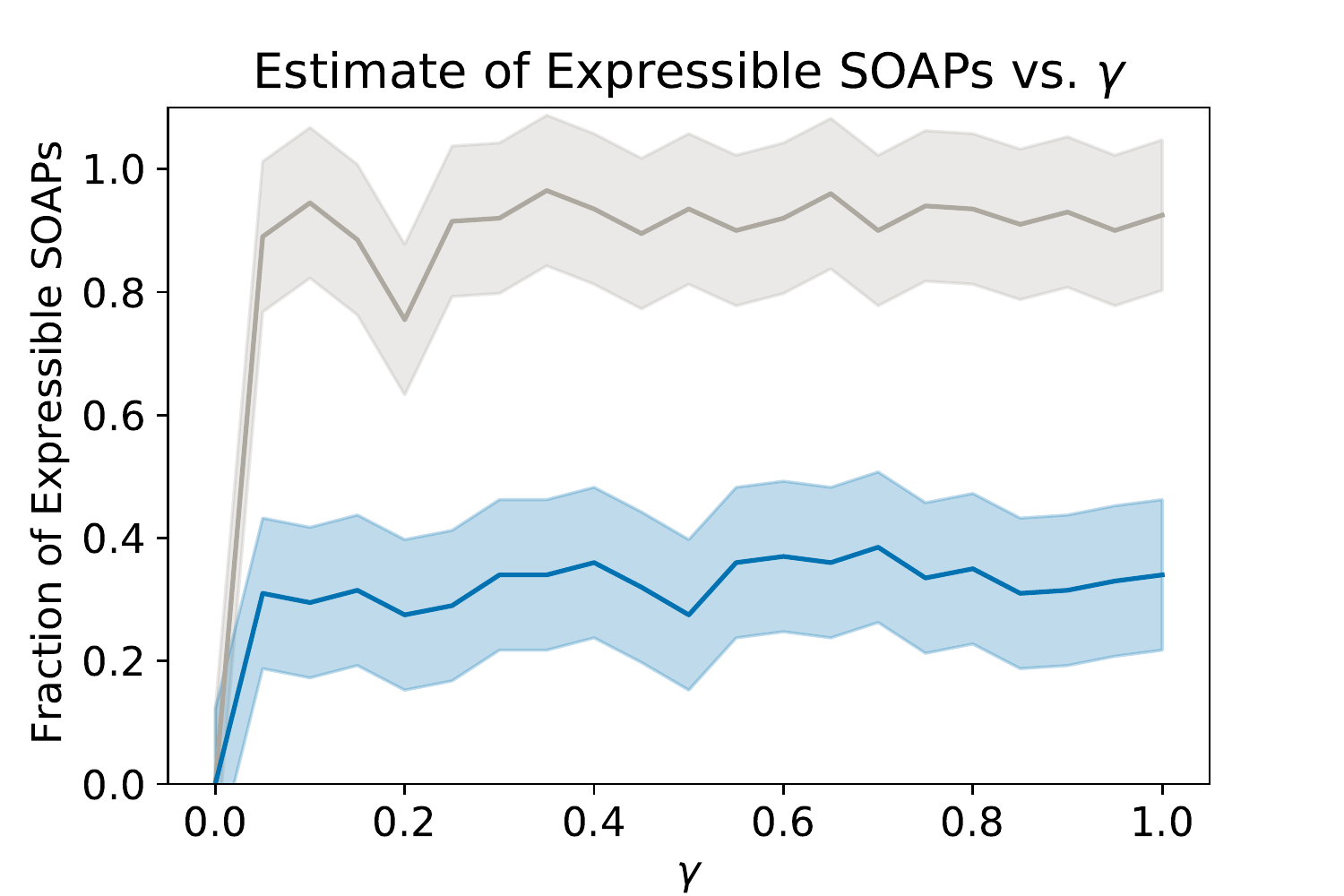}} \\
    %
    \subfloat[Vary SOAP Size\label{subfig:soap_size}]{\includegraphics[width=0.3\textwidth]{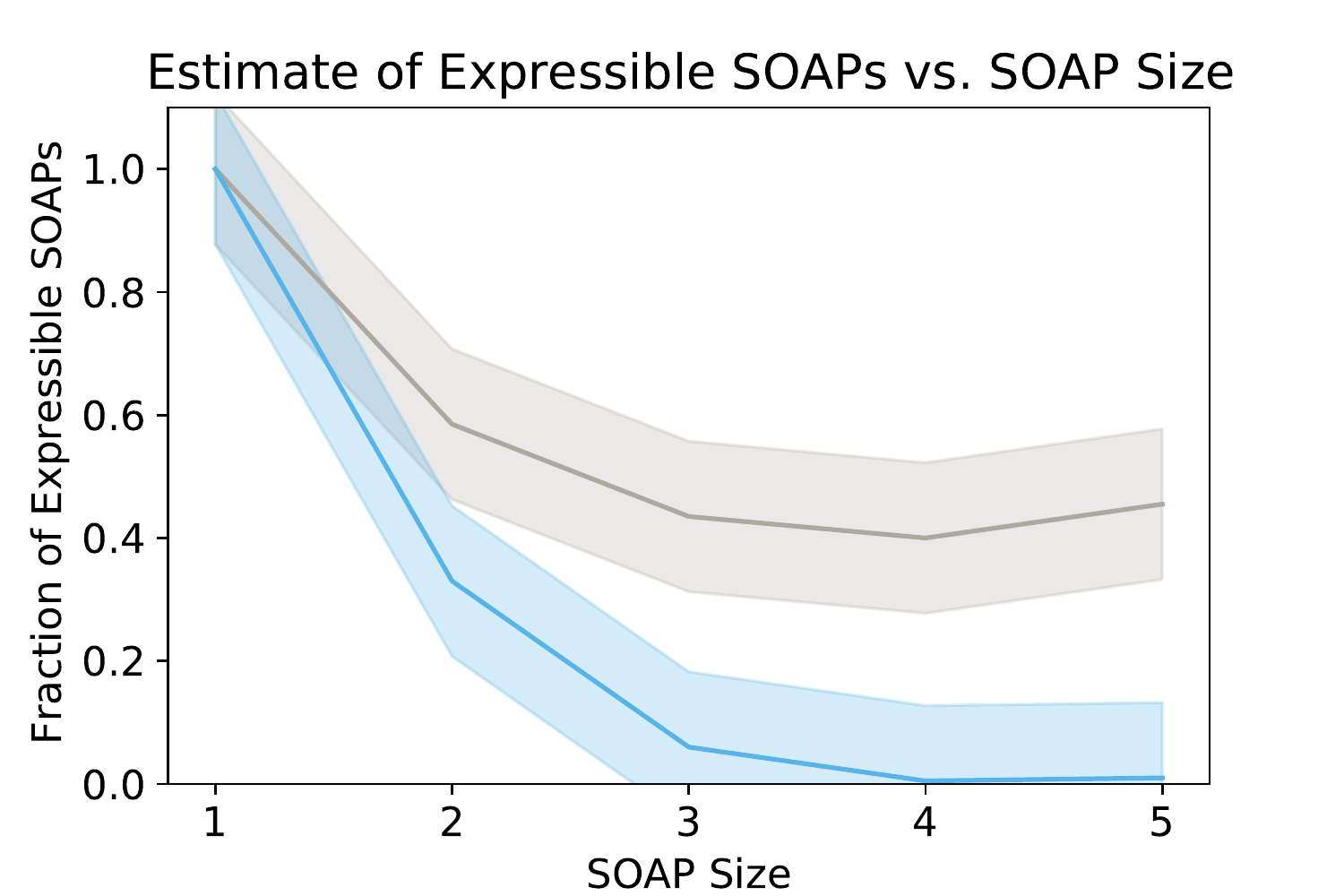}}
    %
    %
    \subfloat[Vary Entropy of $T$\label{subfig:entropy}]{\includegraphics[width=0.3\textwidth]{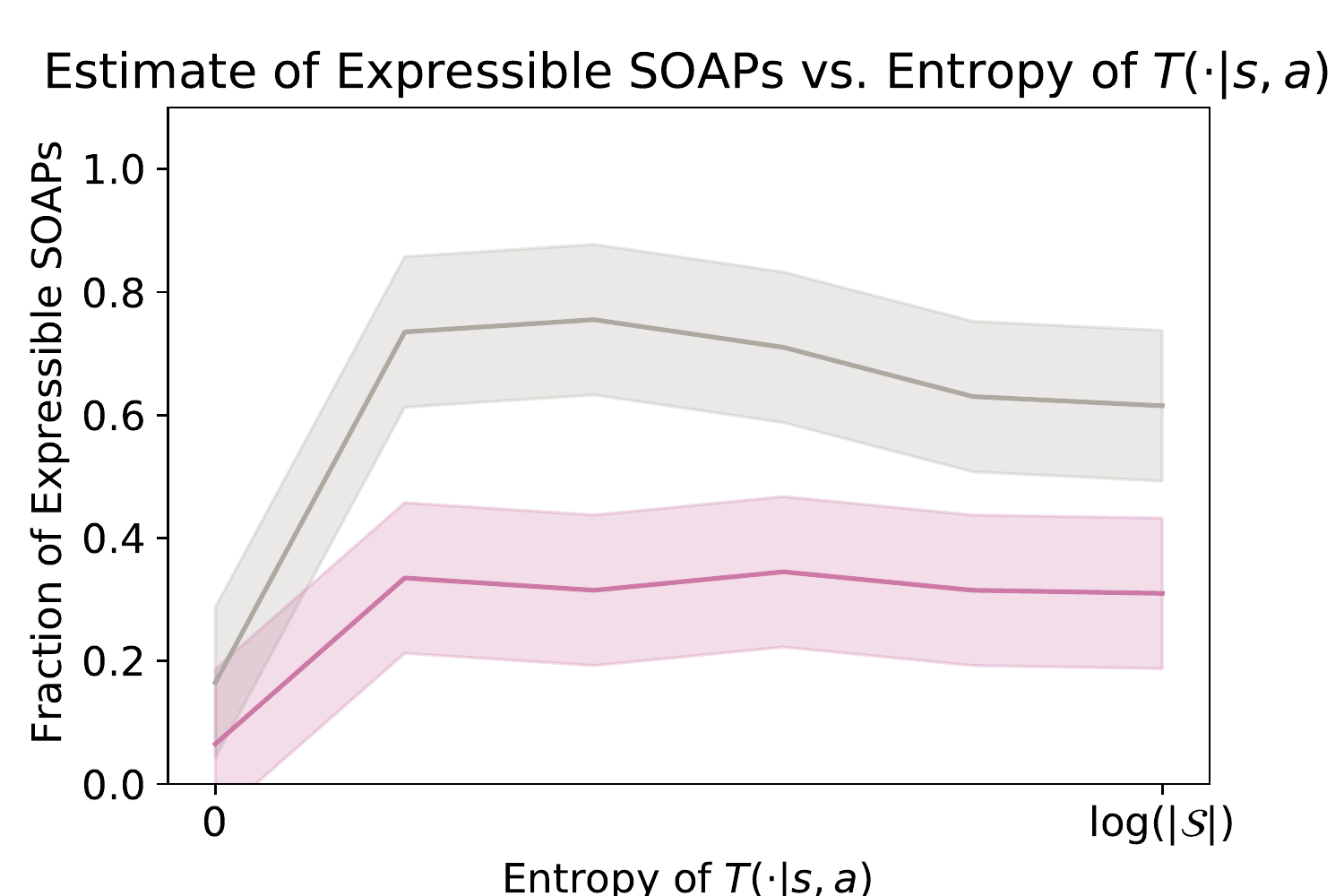}}
    %
    \subfloat[Vary the Spread of $\Pi_G$\label{subfig:spread}]{\includegraphics[width=0.3\textwidth]{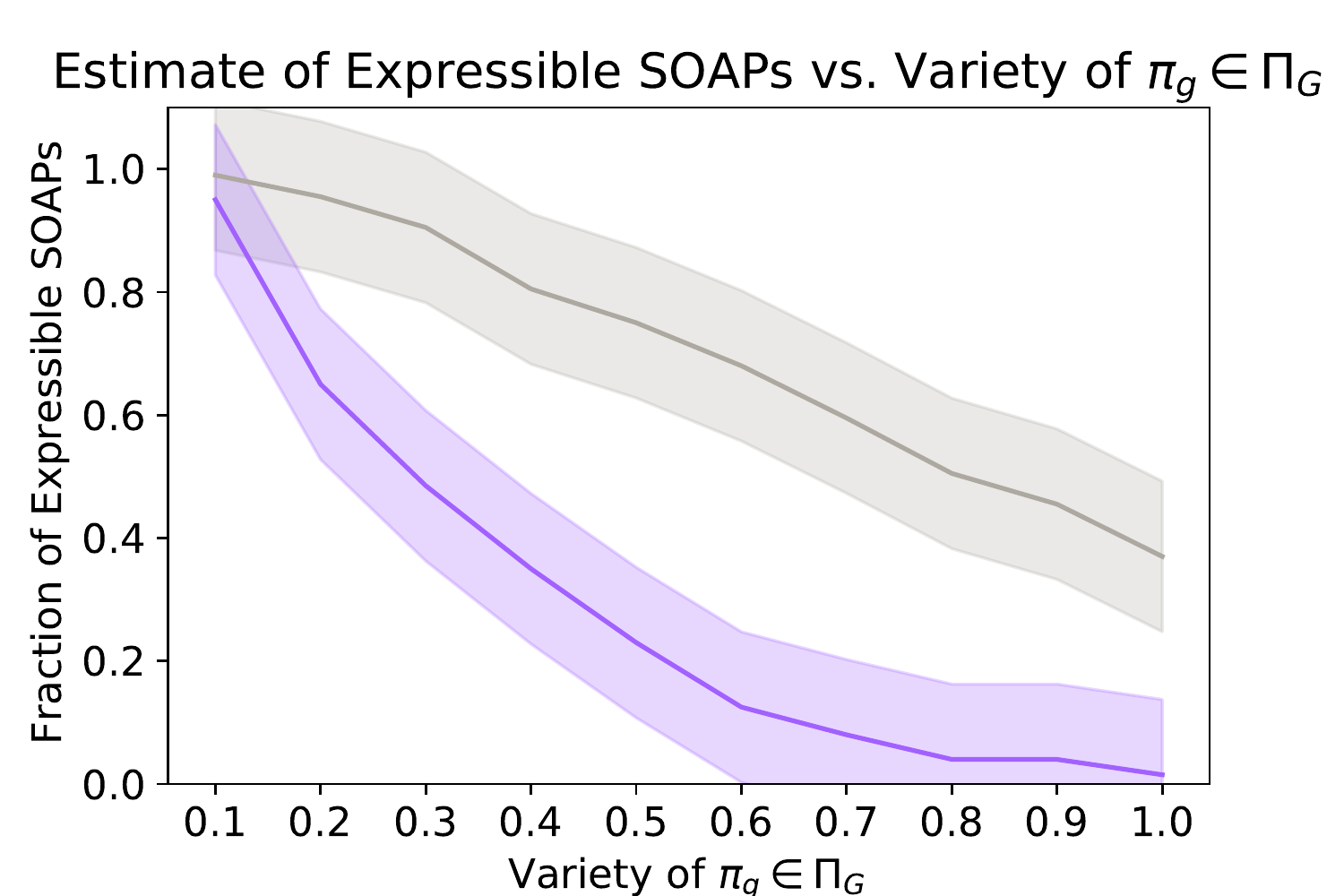}}
    \caption{The approximate fraction of SOAPs that are expressible by reward in CMPs with a handful of states and actions, with 95\% confidence intervals. In each plot, we vary a different parameter of the environment or task to illustrate how this change impacts the expressivity of reward, showing both equal (color) and range (grey) realization of SOAP.}
    \label{fig:results_soap_expressivity}
\end{figure*}

%% file: figures/results_2_learning.tex
\begin{figure*}[t!]
    \centering
    %
    \subfloat[Grid World SOAP Reward\label{subfig:soap_rew_visual}]{\includegraphics[width=0.49\textwidth]{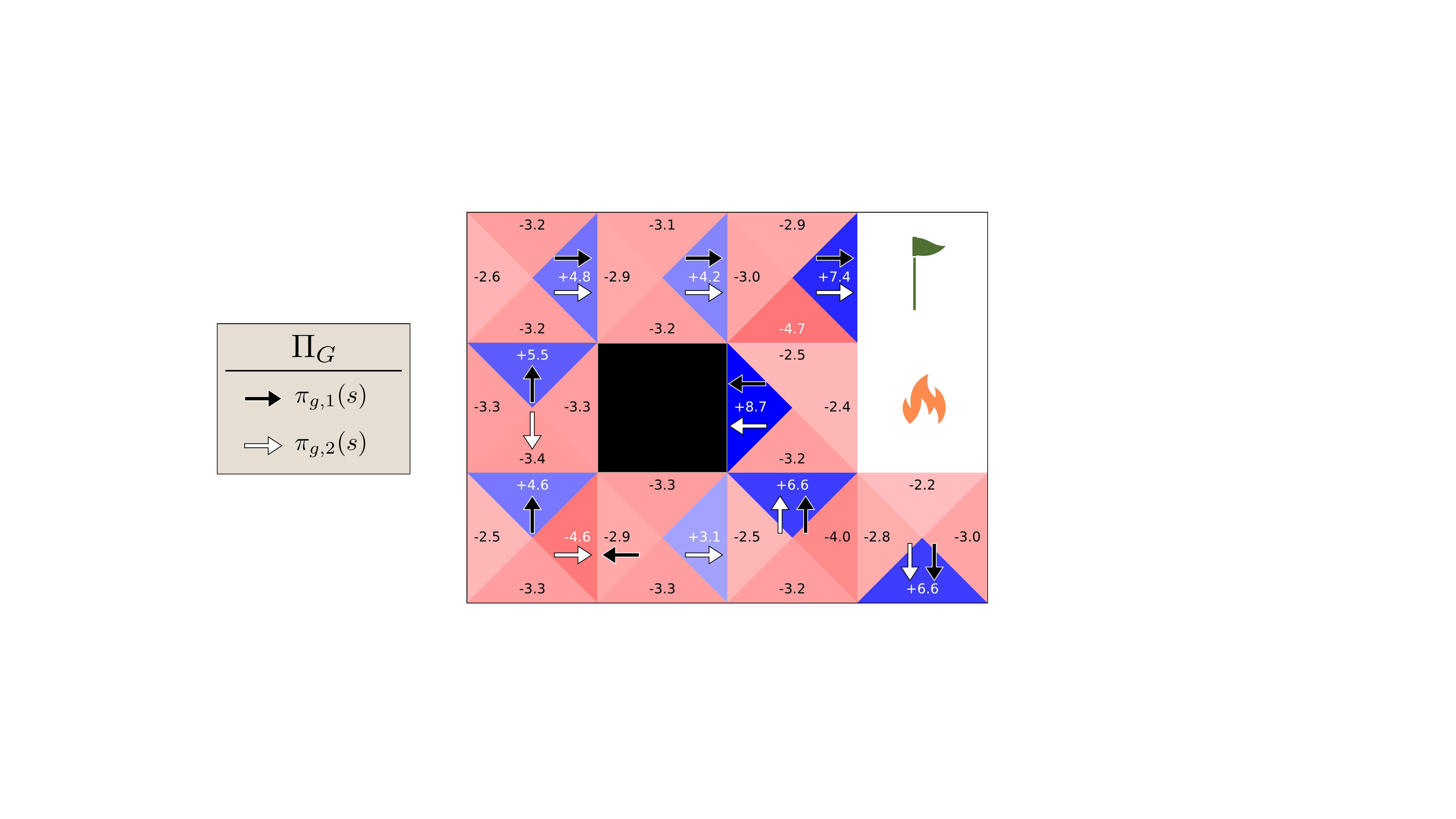}} \hspace{6mm}
    %
    %
    %
    \subfloat[Grid World Learning\label{subfig:grid_learning_curve}]{\includegraphics[width=0.41\textwidth]{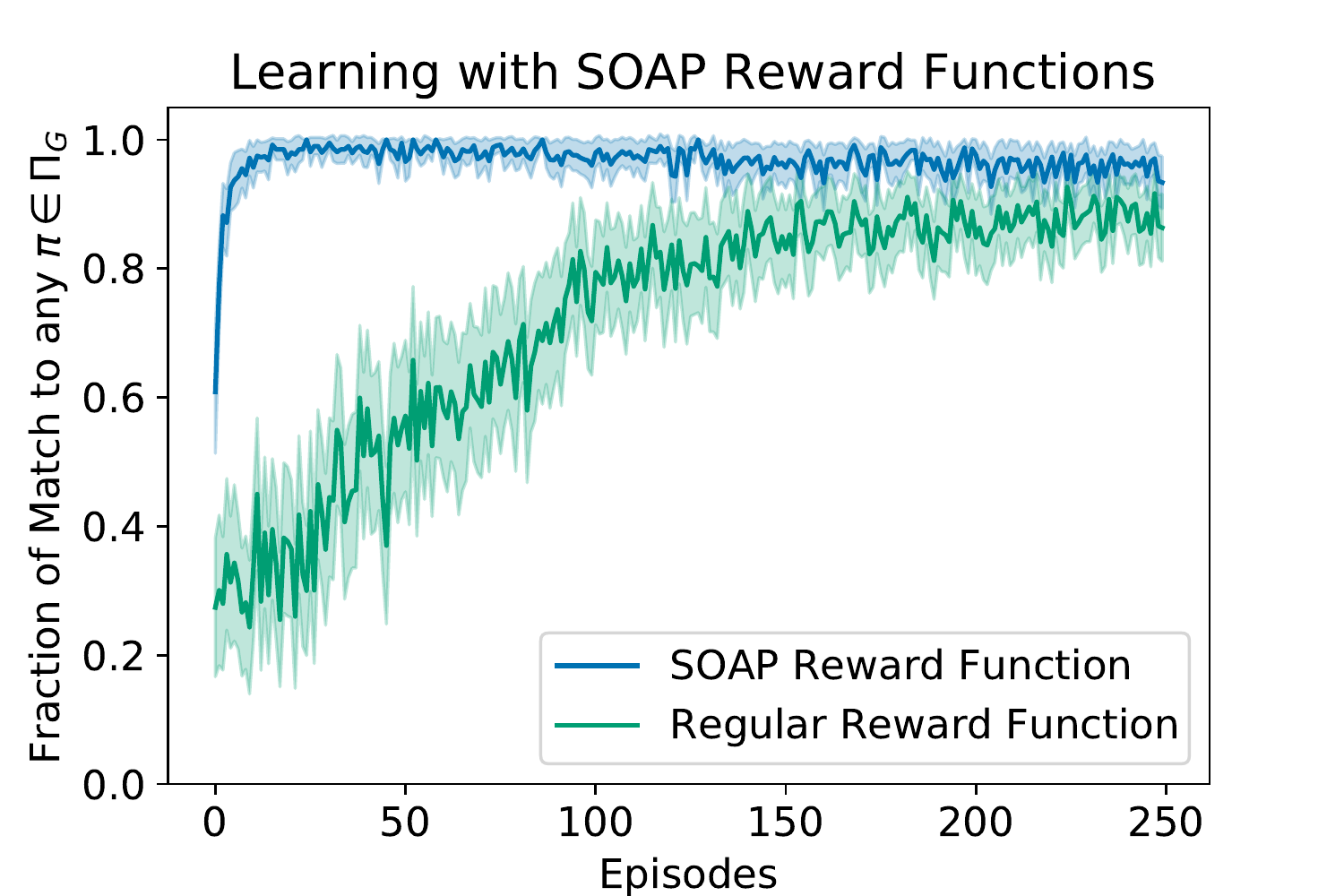}}
    \caption{A SOAP-designed reward function (left) and the resulting learning curves (right) for Q-learning compared to the traditional reward function for the \citet{russell94} grid world. Each series presents average performance over 50 runs of the experiment with 95\% confidence intervals.}
    \label{fig:rn_grid_results}
\end{figure*}

%% file: neurips_appendix-content.tex
\appendix

\section{Anticipated Questions}
\label{apend:anticipated_questions}

We first address questions that might arise in response to the main text.

\begin{enumerate}[label=\textit{(Q\arabic*)}]

    %
    \item \textit{What does it mean for Bob to *solve* one of these tasks? That is, if Alice chooses a SOAP, PO, or TO for Bob to learn to solve, when can Alice determine Bob has solved the task?}
    
    A: Bob can be said to be doing better on a given task if his behavior improves, as is typical in evaluating behavior under reward. The difference with SOAPs, POs, and TOs is that we measure improvement relative to the task rather than reward. For instance, given a SOAP, we might say that Bob has solved the task once he has found one of the good policies, and we might measure Bob's progress on a task in terms of the distance of his greedy policy to one of the good policies (as done in our learning experiments). The same reasoning applies to POs and TOs: Bob is doing better on a task in so far as his greedy policy (or trajectories) is (are) higher up the ordering. 
    
    %
    \item \textit{These notions of inexpressibility all come about due to the Markov restriction on reward functions. That is, the studied reward functions must be a function of $s$, $(s,a)$, or $(s,a,s')$. But, what about history-based reward functions?}
    
    A: Indeed, as discussed in our introduction, our goal is to examine the expressivity of Markov rewards in the context of finite MDPs. We assume the environment is fixed and given to Alice with the state and action spaces already determined. While it is sensible to consider history-based rewards, this opens up new considerations: Must the state space also change so as to retain the Markov property? Instead, we suggest that for a given CMP, it is natural to be interested in Markov rewards, but acknowledge the importance of going beyond such functions. As discussed in the main text, we suspect that there is a coherent account of which tasks are and are not expressible as a consequence of some of the axioms for rationality. We hope to study these directions in future work.
    \vspace{2mm}
    
    %
    \item \textit{Why restrict attention to SOAPs, POs, and TOs?}
    
    A: First, we recognize these do not necessarily capture all of what we hope to convey to learning agents. It is an important next step in our work to enrich the analysis with more general objectives. Still, we believe that these each represent an interesting, relatively general, and concrete template for what a task might look like. They are quite flexible: SOAPs can be simple while POs and TOs can be complex.
    
    %
    \item \textit{Why restrict to the start-state value?}
    
    A: We adopt start-state value due to its simplicity. Other considerations might be: (1) The \textit{expected} value under some chosen distribution, or (2) That the constraint hold over all states (so, for SOAPs, each $\pi_g$ is better than each $\pi_b$ in value for all states). We note that the former case is identical to start-state value, as we can always add a start-state to any CMP where all actions lead to the desired next-state distribution in $T$. The latter case is slightly more complicated, so we chose not to focus on it as we prefer the simplicity of the start-state case. However, we note that many inexpressible tasks under the start-state criterion remain inexpressible under the ``all-state'' criterion (such as the XOR example from Figure 2b).
    
\end{enumerate}


\spacerule
\section{Proofs}
\label{apend:proofs}

We next restate each central result, and present its proof.

\begin{customprop}{3.1}
There exists a CMP, $E$, and choice of $\Pi_G$ such that $\Pi_G$ can be realized under the range-SOAP criterion, but cannot be realized under the equal-SOAP criterion.
\label{prop-a:soap_separation}
\end{customprop}

\begin{figure}[!h]
    \centering
    \includegraphics[width=0.5\textwidth]{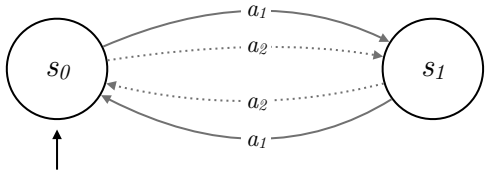}
    \caption{A CMP that separates the two kinds of SOAP realizations.}
    \label{fig:two_soap_example}
\end{figure}

\input{proofs/prop_two_soap_separation}


%
\begin{customthm}{4.1}
For each of SOAP, PO, and TO, there exist $(E,\mathscr{T})$ pairs for which no reward function realizes $\mathscr{T}$ in $E$.
\label{thm-a:main_result_express}
\end{customthm}

\input{proofs/4-1_main_result_proof}

%
\begin{customprop}{4.2}
There exist choices of $E_{\neg T} = (\mc{S}, \mc{A}, \gamma, s_0)$ or $E_{\neg \gamma} = (\mc{S}, \mc{A}, T, s_0)$, together with a task $\mathscr{T}$, such that there is no $(T,R)$ pair that realizes $\mathscr{T}$ in $E_{\neg T}$ or $(R,\gamma)$ in $E_{\neg \gamma}$.
\label{prop-a:fixed_sag_vary_t_r}
\end{customprop}

\input{proofs/4-2_vary_t_inexpr}

%
\begin{customthm}{4.3}
The \textsc{RewardDesign} problem can be solved in polynomial time, for any finite $E$, and any SOAP, PO, or TO, so long as a reward-function family with infinitely many outputs is used.
\label{thm-a:reward_design_is_in_p}
\end{customthm}

\input{proofs/4-3_main_result_proof}


%
\begin{customlem}{B.1}
The SOAP variant of \textsc{RewardDesign} can be solved in polynomial time.
\label{lem-a:soap_rew_design}
\end{customlem}

\input{proofs/reward_design_lemmas/soap_rew_design}

%
\begin{customlem}{B.2}
The PO variant of \textsc{RewardDesign} can be solved in polynomial time.
\label{lem-a:po_rew_design}
\end{customlem}

\input{proofs/reward_design_lemmas/po_rew_design}

%
\begin{customlem}{B.3}
The TO variant of \textsc{RewardDesign} can be solved in polynomial time.
\label{lem-a:to_rew_design}
\end{customlem}

\input{proofs/reward_design_lemmas/to_rew_design}

%
\begin{customcor}{4.4}
For any task $\mathscr{T}$ and environment $E$, deciding whether $\mathscr{T}$ is expressible in $E$ is solvable in polynomial time
\label{cor-a:rew_design_decidable}
\end{customcor}

\input{proofs/4-4_decidability}
%
Next, we provide further details on \autoref{thm-a:finite_po_np-hard} that examines reward design when only finitely many reward outputs may be used. As noted in the main text, \autoref{thm-a:reward_design_is_in_p} requires that Alice is allowed to design a reward function that can produce infinitely many outputs. It is natural to wonder whether this requirement is strict. \autoref{thm-a:finite_po_np-hard} answers this question in the affirmative, by proving that the following decision problem is hard.

%
\begin{definition}
The \textsc{Finite-PO-RewardDecision} problem is defined as follows: \textbf{Given} $E = (\mc{S}, \mc{A}, T, \gamma, s_0)$, and a set of $k$ policy inequalities ($\pi_{x_i} < \pi_{y_i}$), \textbf{output} True iff there is a reward function $R(s')$ that induces the given policy inequalities.
\end{definition}

Note that this formulation focuses on POs, and on reward as a function of next-state. Unfortunately, we find this problem is NP-hard, showing that for reward design to be efficient, infinitely many reward outputs are needed.

%
\begin{customthm}{4.5}
The \textsc{Finite-PO-RewardDecision} problem is NP-hard.
\label{thm-a:finite_po_np-hard}
\end{customthm}

\input{proofs/4-5_finite_np_hard}

\begin{customprop}{4.6}
For any SOAP, PO, or TO, given a finite set of CMPs $\mc{E} = \{E_1, \ldots, E_n\}$ with shared state--action space, there exists a polynomial time algorithm that outputs a single reward function that realizes the task (when possible) in each CMP in $\mc{E}$.
\label{prop-a:multi_environment}
\end{customprop}
\input{proofs/4-6_multi_environment}

%
\begin{customthm}{4.7}
Task realization is not closed under sets of CMPs with shared state-action space. That is, there exist choices of $\mathscr{T}$ and $\mc{E} = \{E_1, \ldots, E_n\}$ such that $\mathscr{T}$ is realizable in each $E_i \in \mc{E}$ independently, but there is not a single reward function that realizes $\mathscr{T}$ in all $E_i \in \mc{E}$ simultaneously.
\label{thm-a:realize_not_closed}
\end{customthm}

\input{proofs/4-7_realize_not_closed}

\begin{figure}[!h]
    \centering
    \includegraphics[width=0.75\textwidth]{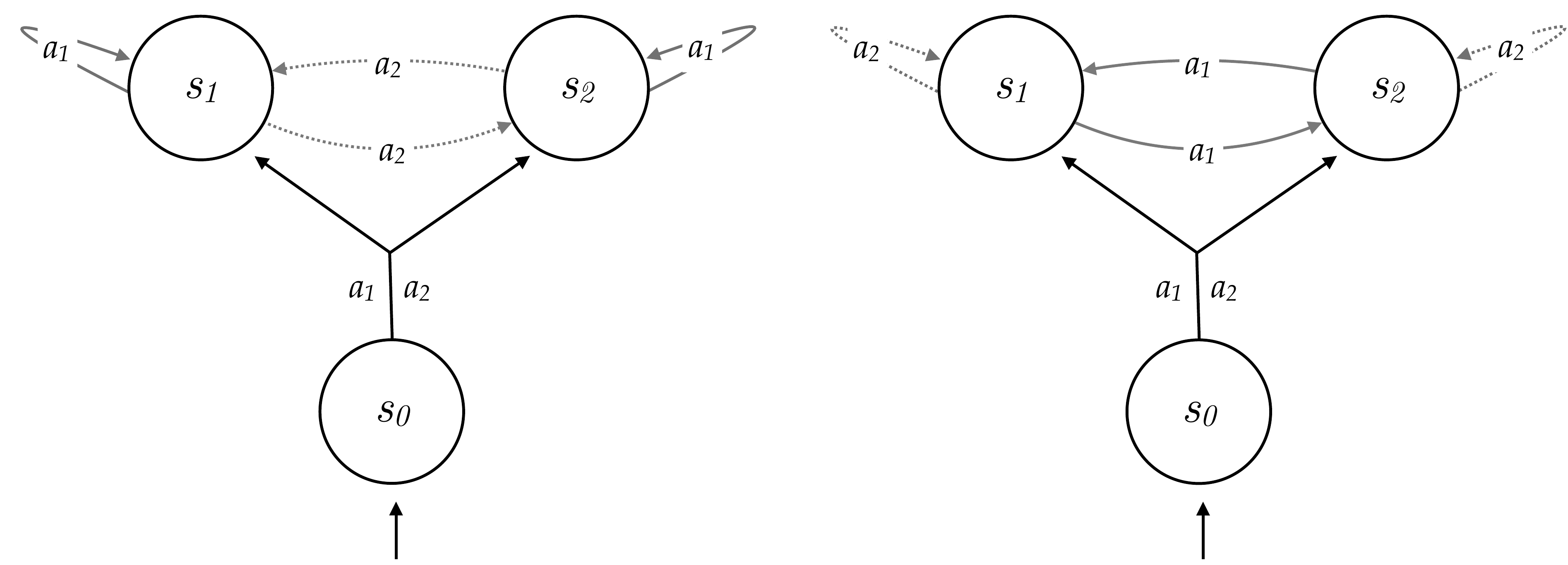}
    \caption{A pair of CMPs with opposite action effects: On the left, $a_1$ keeps the agent in the same place, while $a_2$ flips the state. On the right, the effects are exactly inverted.}
    \label{fig:realization_example}
\end{figure}

\spacerule
\section{Experimental Details}
\label{apend:experiments}

Next, we provide further details about our experiments.

\subsection{Expressibility Experiments}

First, we provide additional information about the first experiment that explores the fraction of SOAPs that are expressible in small CMPs.

\paragraph{Six Variants.} In each figure, we vary one aspect of the environment or task along the x-axis. Most of these are self-explanatory (a: number of actions, b: number of states, c: $\gamma$), though the plots in (d), (e) and (f) are slightly more involved. In (d), we vary the \textit{size} of each sampled SOAP, corresponding to the number of good policies in the SOAP. That is, a SOAP of size one consists only of a single good policy, $\Pi_G = \{\pi^*\}$. In (e), we vary the \textit{Shannon entropy} of the transition function on a per state-action basis as per: $H(T) = -\sum_{s' \in \mc{S}} \log_2 T(s' \mid s,a)$. This is accomplished by interpolating between the fully deterministic transition function that only transitions to a single next-state and the uniform random transition function through a simple soft-max distribution in which one next-state is the intended transition, while each other next-state receives a small amount of probability mass depending on the given entropy. In plot (f), we vary the \textit{spread} of the SOAP, which is a measure of how different the good policies in the SOAP are on average. The x-axis corresponds to an approximate edit-distance between policies, where each point on the x-axis defines the parameter of a coin that we flip to determine whether to change a chosen reference policy's action for each state. So, we first randomly sample one policy, say $\pi_1$. Then, we construct the next policy for the SOAP as follows: For a given coin weight $\theta$, we flip a coin at each state of the CMP. If the coin lands heads (the trial is successful), then we change the action of the new policy to a fixed action chosen uniformly at random. Thus, when $\theta$ is zero, the SOAP will only contain $\pi_1$. When $\theta$ approaches one, the SOAP will likely contain many different policies.

%
\paragraph{Environment Details.} In each case, unless otherwise specified, the underlying environment is a four state, three action CMP with $\gamma=0.95$, and a transition function that is a multinomial over next-states sampled from a Dirichilet-multinomial distribution with $\alpha$ parameters set to $\frac{1}{|\mc{S}|}$. When not specified, the size of each SOAP is two. We varied many aspects of these parameters and found little change in the nature of the plots, though trends will be shifted up or down in many cases. For instance, given the downward trend of Figure (3d) as the SOAP size increases, we know that the remaining plots will each be scaled downward if we were to run the same experiment for a SOAP size larger than two. We sample random SOAPs by first sampling a SOAP size randomly between $1$ and $|\Pi|$. Then, we sample $N$ SOAPS of the chosen size uniformly at random (unless otherwise specified, as in the case of Figure (3f)).

%
\subsection{Learning Experiments}

In the grid environment, we set slip probability of 0.35 for all (non-terminal) states. When a ``slip'' event occurs, the action effect is orthogonal to the intended direction. For instance, in the bottom left cell, if the \texttt{up} action is executed, there is a 0.175 chance the agent will execute \texttt{left} (thus staying in the bottom right cell), and a 0.175 chance the agent will execute \texttt{right}, and a 0.65 chance the agent will move up a cell. We experiment with tabular Q-learning with $\epsilon$-greedy exploration, with $\epsilon=0.2$ and learning rate $\alpha=0.1$ and no annealing. Each episode consists of 10 steps in the environment, with 250 episodes per algorithm. We repeat the experiment 50 times and report 95\% confidence intervals. The y-axis measures, at the end of each episode, the (inverse) minimum edit distance between Q-learning's greedy policy and any of the policies in the SOAP along the trajectory taken by Q-learning's greedy policy. Thus, when the series reaches 1.0, Q-learning's greedy policy is identical to one of the SOAP policies in the states that the greedy policy will reach. We observe that the gap between the blue and green curves is due to the different kinds of policies that the SOAP reward and the regular reward promote---one is not necessarily better or worse than the other, they just convey different kinds of objectives.

\spacerule

%% file: proofs/prop_two_soap_separation.tex
\begin{dproof}[\autoref{prop-a:soap_separation}]
Consider the example in \autoref{fig:two_soap_example} and the SOAP $\Pi_G = \{\pi_{11}, \pi_{21}, \pi_{12}\}$. This $\Pi_G$ indicates that all policies are acceptable except the policy that always takes $a_2$. That is, the policy subscripts denote which actions each policy takes in each state ($\pi_{12}$ means $a_1$ in $s_0$, $a_2$ in $s_1$).

First, let us note that range SOAP is realizable: The listed rewards allow for each policy in $\Pi_G$ to obtain $\frac{\textsc{RMax}/2}{1-\gamma}$ value or better, while $\pi_{22}$ achieves zero.  Letting $\eps = \textsc{Vmax}/2$, this choice of rewards satisfies the criteria, and the $\Pi_G$ is $\eps$-realized in the given MDP.

Next, note that there can exist no other choice of rewards that realize the equal SOAP. That is, such that $V^{\pi_{11}}(s_0) = V^{\pi_{12}}(s_0) = V^{\pi_{21}}(s_0) > V^{\pi_{22}}(s_0)$. This fact is a consequence of the tie in values between the policies. Here, we see that any choice of rewards that makes $V^{\pi_{12}}(s_0) = V^{\pi_{21}}(s_0)$  will also give $\pi_{22}$ that same value. Thus, the given $\Pi_G$ is unrealizable under equal SOAP. \qedhere

\end{dproof}

%% file: proofs/4-1_main_result_proof.tex
\begin{dproof}[\autoref{thm-a:main_result_express}]

We proceed by proving the existence of a pair ($E, \mathscr{T}$), for each of $\mathscr{T}$ as a SOAP, PO, or TO. Indeed, we find that the simple XOR case is inexpressible for all three task types.

\textbf{SOAP.} For SOAP, we consider the CMP with two states and two actions from \autoref{fig:two_soap_example}, and the SOAP $\Pi_G = \{\pi_{12}, \pi_{21}\}$. That is, the chosen task is for the learning agent to find a policy that chooses each action in exactly one state. Here, we find that any Markov reward function that makes $a_1$ optimal in the left state will, by consequence, make $a_1$ an optimal action no matter what is done in other states. In other words, we cannot assign rewards to $(s,a)$ pairs so that an action's optimality depends on \textit{which} optimal action is taken in the other state. Thus, all choices of Markov reward function that make $\{\pi_{12}, \pi_{21}\}$ optimal will also make $\{\pi_{11}, \pi_{22}\}$ optimal, too.

\textbf{PO.} Since the given SOAP is a special case of a PO, we have already identified a given inexpressible PO.

\textbf{TO.} For TO, for simplicity we consider the same CMP. We let $N=2$, and suppose that the desired trajectory ordering is over state-action pairs, giving rise to a set of good trajectories, and a set of bad trajectories:
\begin{align}
    L_{\tau, N} &:= \{\tau_G, \tau_B\}, \\
    &\tau_G = \{\left\{(s_0, a_1), (s_1, a_2)\}, \{(s_0, a_2), (s_1, a_1)\}\right\}, \\
    &\tau_B = \{\left\{(s_0, a_1), (s_1, a_1)\}, \{(s_0, a_2), (s_1, a_2)\}\right\}.
\end{align}
The same reasoning from the above cases applies: We cannot make the good trajectories strictly higher in return than the bad trajectories.\qedhere
\end{dproof}

%% file: proofs/4-2_vary_t_inexpr.tex
\begin{dproof}[\autoref{prop-a:fixed_sag_vary_t_r}]

The running XOR example is actually inexpressible for \textit{all} choices of $T$, or of $\gamma$. That is, there is no way to make $\pi_{12}$ and $\pi_{21}$ strictly better than both $\pi_{22}$ \textit{and} $\pi_{11}$ by varying $\gamma$ or $T$. Such examples likely exist for any choice of $\mc{S}$ and $\mc{A}$ of size greater than one. \qedhere
\end{dproof}

%% file: proofs/4-3_main_result_proof.tex
\begin{dproof}[\autoref{thm-a:reward_design_is_in_p}]

We proceed by providing constructive algorithms for each of SOAP, PO, or TO. All three are based on similar applications of a linear program (LP), though there is nuance that separates them. We present each as a Lemma (\autoref{lem-a:soap_rew_design}, \autoref{lem-a:po_rew_design}, \autoref{lem-a:to_rew_design}), which together constitute the proof of this Theorem. \qedhere
\end{dproof}

%% file: proofs/reward_design_lemmas/soap_rew_design.tex
\begin{dproof}[\autoref{lem-a:soap_rew_design}]
  
We proceed by constructing a linear program (LP) whose solution is the desired reward function (or correctly outputs that there is no such reward function). Specifically, note that we want to choose a reward function so that all the policies in $\Pi_G$ have strictly higher start-state value than all the policies not in the set. We present the proof through five observations.

%
\paragraph{First,} observe that any reward function that will induce the desired ordering ensures that the optimal policy $\pi^*$ is in the set $\Pi_G$. This is true since $\pi^*$ (under the chosen reward function) is better than all policies. So it is better than all policies not in $\Pi_G$.

%
\paragraph{Second,} note that the set $\Pi_G$ is well connected in the following sense. Let a step in policy space from some reference policy $\pi_{\text{ref}}$ to be a move to any other deterministic policy that differs from $\pi_{\text{ref}}$ in exactly one state.  Then, for any pair of policies in $\Pi_G$, there must be a sequence of policies in $\Pi_G$, each one step apart from the next, from one to the other. This follows from the policy-improvement theorem: we can get from any policy to an optimal policy in a sequence of policies such that each policy (1) is one step from the previous one and (2) strictly dominates the previous one. Since any policy that strictly dominates a policy in $\Pi_G$ must be better than the policy in $\Pi_G$, it must also be in $\Pi_G$ (if the problem constraint is satisfied). That means if we choose two policies in $\Pi_G$, $\pi_1$ and $\pi_2$, both can reach $\pi^*$ in a sequence of single steps while staying within $\Pi_G$. Since steps are symmetric, $\Pi_G$ is connected.
The connected set of policies in $\Pi_G$ has a ``fringe'' $\Pi_{\text{fringe}}$---a set of policies not in $\Pi_G$ that are one step from a policy in $\Pi_G$.

%
\paragraph{Third,} for the constraints of the problem to be satisfied, every policy $\pi_g \in \Pi_G$ must be strictly better than every policy $\pi_f \in \Pi_{\text{fringe}}$.

%
\paragraph{Fourth,} observe that $|\Pi_{\text{fringe}}| <= |\mc{A}| |\Pi_G|$, so $\Pi_{\text{fringe}}$ is polynomial sized.

%
\paragraph{Fifth,} we can construct a polynomial-sized LP that expresses that every policy $\pi_g \in \Pi_G$ is strictly better than every policy $\pi_f \in \Pi_{\text{fringe}}$. Note that the direct way to build this LP has a “strictly better than” comparison between each policy $\pi_f \in \Pi_{\text{fringe}}$ and each policy $\pi \in \Pi_G$. That’s at most $|\mc{A}| |\Pi_G|^2$ inequalities.

%
We now tie the above observations together to show that the solution to this LP solves the constraints of the problem. That is, the reward function returned makes it so every policy $\pi_g \in \Pi_G$ is strictly better than every policy not in $\Pi_G$, \textit{and} no valid reward function is excluded (so, if a solution exists, it will be found).

%
The argument that no valid reward function is excluded is simply because the set of constraints in the LP is a subset of the defining constraints of the problem. Specifically, the LP constrains the policies inside $\Pi_G$ to be strictly better than the ones on the fringe instead of all policies not in $\Pi_G$.


%
The argument that only constraining the values on the fringe automatically constrains all the others proceeds as follows. First, with respect to the returned reward function, there is some optimal policy $\pi^*$. That policy $\pi^*$ must be in $\Pi_G$. To see why, let us assume it is not. That means there is a sequence of improving steps that turn a policy in $\Pi_G$ to $\pi^*$ (currently assumed to be out of $\Pi_G$). But, any such sequence must go through the fringe, and we constrained the fringe so that all of the policies in $\Pi_G$ are strictly better than them. So, $\pi^*$ must be in $\Pi_G$. Next, we know that all policies in $\Pi_G$ must be strictly better than the policies not in $\Pi_G$. To see why, consider an ``improving'' path from some policy $\pi_b \not \in \Pi_G$ to $\pi^*$. Since $\pi^*$ is in $\Pi_G$, we know this path must go through some policy $\pi_f \in \Pi_{\text{fringe}}$. Since it’s an improving path, that means $\pi_f$ is better than $\pi_b$. But, every policy in $\Pi_G$ is strictly better than $\pi_f$, so it must also be strictly better than $\pi_b$. \qedhere
\end{dproof}

%% file: proofs/reward_design_lemmas/po_rew_design.tex
\begin{dproof}[\autoref{lem-a:po_rew_design}]


We proceed by constructing a procedure that calls a linear program whose answer is the  reward function that induces the given $L_\Pi$ in $M$, or the procedure correctly outputs that there is no such $R$.

Consider the set of policies in $\Pi$, numbered $\pi_1, \ldots, \pi_i, \ldots, \pi_N$. Note that the value of $\pi_i$ in $s_0$ can be computed in terms of the expected reward under the policy's discounted expected state-action visitation distribution. That is, for each $\pi_i$, let
\begin{equation}
    \rho_i(s,a) := \sum_{t=0}^\infty \gamma^t \Pr(s_t = s, a_t = a \mid s_0, \pi_i).
\end{equation}
Since the given MDP is assumed to have finite state-action space, note that $\rho_i$ may be interpreted as a vector whose elements correspond to $\rho_i(s_0,a_0), \rho_i(s_0,a_1)$, and so on.

The value of $\pi_i$ under a given reward function $R$ (which may also be interpreted as a vector) is then produced by the dot product $R \cdot \rho_i$.

Given $L_\Pi$, we want to find an $R$ that ensures a set of linear constraints hold:
\begin{equation}
    R \cdot \rho_0 \geq R \cdot \rho_1 \geq \ldots.
\end{equation}

Note that the trivial reward function, $R_{\emptyset} : s \mapsto 0$, is a solution to the above linear program. However, we can ensure some minimal increment improvement of $\epsilon$ for non-tying policies, where
\begin{equation}
    R \cdot \rho_0 \geq R \cdot \rho_1 + \epsilon \geq \ldots.
\end{equation}
This $\epsilon$ minimal increment is sufficient to separate policies with tying scores and avoids the degenerate solution of $R_\emptyset$, so long as there are infinitely many reward outputs feasible.



%
Note that the input is of size $N$, where $N$ is the number of constraints imposed on the policy ordering. In the worst case, $N = |L_\Pi| \leq |\mc{A}|^{|\mc{S}|}$. If there are fewer constraints than either $|\mc{A}|$ or $|\mc{S}|$, then $N = \max\{|\mc{S}|, |\mc{A}|\}$. The amount of computational work required is split across two steps:
\begin{enumerate}
    \item $\tilde{O}(N)$: Compute $\rho_i$ for each policy $\pi_1 \ldots \pi_N$. 
    \item ${O}(N^3)$: Formulate and solve the above linear program.
\end{enumerate}

Thus, since the described linear program can be constructed in polynomial time outputs a reward function that induces the given $L_\Pi$, we conclude the PO case. \qedhere
\end{dproof}

%% file: proofs/reward_design_lemmas/to_rew_design.tex
\begin{dproof}[\autoref{lem-a:to_rew_design}]
The algorithm follows the same construction as those catered toward SOAPs and POs, but is in fact much simpler.

We can form linear inequality constraints on the return of two trajectories as follows. First recall that the $N$-step discounted return of a trajectory $\tau$ is
\begin{equation}
    G(\tau;s_0) = \sum_{i=0}^{N-1} \gamma^i R(s_i, a_i),
\end{equation}
assuming reward is a function of state and action for simplicity. Note that because reward functions are assumed to be deterministic, the quantity $G(\tau;s_0)$ is not a random variable.

Now, given two trajectories,
\begin{equation}
    \tau_i = \{(s_0^{(i)},a_0^{(i)}), \ldots, (s_{N-1}^{(i)}, a_{N-1}^{(i)})\},\hspace{4mm} \tau_j = \{(s_0^{(j)},a_0^{(j)}), \ldots, (s_{N-1}^{(j)}, a_{N-1}^{(j)})\},
\end{equation}
note that we can express linear inequality constraints as follows,
\begin{equation}
\tau_i \cdot R - \tau_j \cdot R \leq \epsilon,    
\end{equation}
where $R$ is a length $N$ state-action vector, and $\epsilon \in \mathbb{Q}_{\geq 0}$ is a slack variable to be maximized as part of the optimization. Inequality constraints follow the same structure, only simpler. By the same reasoning that underlies the construction of the SOAP and PO based algorithms, the above set of constraints define a linear program whose solution is the realizing reward function, if it exists. \qedhere
\end{dproof}

%% file: proofs/4-4_decidability.tex
\begin{dproof}[\autoref{cor-a:rew_design_decidable}]
For each of PO and TO, the constraints we construct define precisely the space of constraints that constitute the given task. Thus, since linear programming will find a solution for the given constraint set, we know that these two forms of the algorithm will also correctly decide when no reward function exists.

The SOAP case is slightly more involved, but still relatively straightforward. We note that the policy fringe, $\Pi_{\text{fringe}}$, is a subset of $\Pi_B$, since $\Pi_{\text{fringe}}$ consists of some policies \textit{not} in $\Pi_G$ by construction. This means that the constraints produced that separate each good policy from a fringe policy in value are a subset of the true constraints (those that separate each $\pi_g \in \Pi_G$ from each $\pi_b \in \Pi_B$). Hence, since constraint relaxations of this kind have the property that they do not exclude solutions, we conclude that our proposed linear program will correctly determine when no satisfying reward function exists. \qedhere
\end{dproof}

%% file: proofs/4-5_finite_np_hard.tex
\begin{dproof}[\autoref{thm-a:finite_po_np-hard}]


We assume every $T(s' \mid s,a)$ is expressed as a rational number. We also assume that all policies are deterministic, Markov policies, although results should extend to stochastic policies with rational probabilities as well.

The binary PO problem is the same, but it insists that every $R(s') \in \{0,1\}$ for all $s'$ in the returned reward function.


\textbf{Observation 1:} The binary PO decision problem is in NP. We can guess an assignment of $R(s,a)$ to either 0 or 1, then evaluate in inequality using linear equation solving as policy evaluation.

We show that the binary PO decision problem can be used to decide the NP-hard monotone clause 3-SAT problem with a polynomial reduction.

A monotone clause 3-SAT problem consists of a set of $n$ variables, and $m$ clauses. Each clause consists of three variables and is either a positive clause or a negative clause. In a positive clause, all three variables appear as literals. In a negative clause, all three variables appear as negated literals. The problem is the same as the standard 3-SAT problem except for the restriction that we cannot mix positive and negative literals in a clause.

We can convert an instance of monotone clause 3-SAT to the binary PO problem as follows. There is only one state where decisions are possible. It is the initial state of the MDP. Each action from this state results in an action-specific probabilistic transition to a set of terminal states, each of which is associated with a terminal reward value.

Because of the simple structure of this MDP, each policy corresponds to an action and vice versa. And, each terminal state corresponds to a reward and vice versa. So, each policy can be viewed as a convex combination of rewards.

We create two terminal states $s_0$ and $s_1$ and create one action ($a_0$) that transitions directly to $s_0$ and one ($a_1$) that transitions directly to $s_1$. We then add a policy constraint that says the $a_0 < a_1$. Because all rewards are in $\{0,1\}$, that forces the reward for $s_0$ to be 0  and for $s_1$ to be 1. Those become our logical primitives, in a sense.

Next, we add $2n$ states, one for each positive and negative literal in the 3-SAT problem. For each variable $v$, we add an action $a_v$ with a 50-50 transition to $s_v$ and $s_{\overline{v}}$, along with two constraints: $a_0 < a_v < a_1$. These constraints ensure that the reward assignment to $s_v$ and $s_{\overline{v}}$ can be interpreted as an assignment to the literals where one gets a 1 and the other gets a zero. There is no other way to satisfy these constraints.

Now, for each positive clause $c$ consisting of variables $v_1$, $v_2$, and $v_3$, we create an action $a_c$ that transitions to $s_{v_1}$, $s_{v_2}$, and $s_{v_3}$ with equal probability. We add a policy constraint that $a_c > a_0$, forcing the assignment of rewards to the variables to correspond to a satisfying assignment for that clause. (At least one of the rewards needs to be set to 1.)

For the each negative clause $c$ consisting of variables $\overline{v_1}$, $\overline{v_2}$, and $\overline{v_3}$, we create an action $a_c$ that transitions to $s_{\overline{v_1}}$, $s_{\overline{v_2}}$, and $s_{\overline{v_3}}$ with equal probability. We add a policy constraint that $a_c < a_1$, forcing the assignment of rewards to the variables to correspond to a satisfying assignment for that clause. (At least one of the rewards needs to be set to 0.)

By the way the MDP is constructed, the constraints are satisfied if and only if the rewards represent a satisfying assignments for the given monotone clause 3-SAT formula. Therefore, an efficient solution to the binary PO decision problem would provide an efficient solution to the NP-hard monotone clause 3-SAT problem. \qedhere
\end{dproof}

%% file: proofs/4-6_multi_environment.tex
\begin{dproof}[\autoref{prop-a:multi_environment}]
From \autoref{thm-a:reward_design_is_in_p}, we know that there is an algorithm to solve the reward design problem for any task and a single environment, in polynomial time. We form the multi-environment algorithm by simply combining the constraints formed by each individual linear program. By the properties of linear programming, the resulting solution will either satisfy \textit{all} of the given constraints, as desired, or will correctly identify that no such satisficing solution exists. \qedhere
\end{dproof}

%% file: proofs/4-7_realize_not_closed.tex
\begin{dproof}[\autoref{thm-a:realize_not_closed}]
We consider a pair of CMPs, $(E_X, E_Y)$, with the same three states and two actions. We will show that there exists choice of $\mathscr{T}$ such that $\mathscr{T}$ is realizable in $E_X$ and $E_Y$ independently, but \textit{not} in both simultaneously. Our result assumes we restrict to reward functions that are only a function of state, but we suspect similar cases exist for reward functions on $(s,a)$ pairs and $(s,a,s')$ triples.

Consider the pair of CMPs in \autoref{fig:realization_example}. These two CMPs share a state-action space and start-state, but not a transition function (and, say, a $\gamma > 0.5$). Let us further suppose we are interested in the SOAP $\Pi_G = \{\pi_{112}, \pi_{212}\}$, that is, the policies that take $a_1$ in $s_1$, and $a_2$ in $s_2$. In both CMPs, the transition function from $s_0$ transitions to $s_1$ with probability 0.5 and $s_2$ with probability 0.5, for both actions.

We first show that this $\Pi_G$ is realizable in both CMPs. For the CMP on the left, note that the reward function $R(s_1) = 1, R(s_2) = -1$, with $\gamma = 0.95$ will ensure $V^{\pi_{112}}(s_0) = V^{\pi_{212}}(s_0)$, and that both policies are strictly better than all others. Next, note that the same is true for the example on the right where $R(s_1) = -1, R(s_2) = 1$. Thus, $\Pi_G$ is independently realizable in both of these CMPs.

However, there cannot exist a reward function that makes $\pi_{112}$ and $\pi_{212}$ strictly better than all other policies in both CMPs---it is either better to stay in $s_1$, or to stay in $s_2$, but it cannot be the case that both are true simultaneously. \qedhere
\end{dproof}

%% file: main.bbl
\begin{thebibliography}{66}
\providecommand{\natexlab}[1]{#1}
\providecommand{\url}[1]{\texttt{#1}}
\expandafter\ifx\csname urlstyle\endcsname\relax
  \providecommand{\doi}[1]{doi: #1}\else
  \providecommand{\doi}{doi: \begingroup \urlstyle{rm}\Url}\fi

\bibitem[Abbeel and Ng.(2004)]{abbeel2004apprenticeship}
Pieter Abbeel and Andrew~Y. Ng.
\newblock Apprenticeship learning via inverse reinforcement learning.
\newblock In \emph{Proceedings of the International Conference on Machine
  learning}, 2004.

\bibitem[Ackley and Littman(1992)]{ackley1992interactions}
David Ackley and Michael~L. Littman.
\newblock Interactions between learning and evolution.
\newblock \emph{Artificial Life II}, 1992.

\bibitem[Akshay et~al.(2013)Akshay, Bertrand, Haddad, and
  Helouet]{akshay2013steady}
Sundararaman Akshay, Nathalie Bertrand, Serge Haddad, and Loic Helouet.
\newblock The steady-state control problem for {M}arkov decision processes.
\newblock In \emph{Proceedings of the International Conference on Quantitative
  Evaluation of Systems}, 2013.

\bibitem[Amin et~al.(2017)Amin, Jiang, and Singh]{amin2017repeated}
Kareem Amin, Nan Jiang, and Satinder Singh.
\newblock Repeated inverse reinforcement learning.
\newblock In \emph{Advances in Neural Information Processing Systems}, 2017.

\bibitem[Bellemare et~al.(2017)Bellemare, Dabney, and
  Munos]{bellemare2017distributional}
Marc~G. Bellemare, Will Dabney, and R{\'e}mi Munos.
\newblock A distributional perspective on reinforcement learning.
\newblock In \emph{Proceedings of the International Conference on Machine
  Learning}, 2017.

\bibitem[Christian(2021)]{christian2021alignment}
Brian Christian.
\newblock \emph{The Alignment Problem: Machine Learning and Human Values},
  pages 130--131.
\newblock Atlantic Books, 2021.

\bibitem[Christiano et~al.(2017)Christiano, Leike, Brown, Martic, Legg, and
  Amodei]{christiano2017deep}
Paul~F. Christiano, Jan Leike, Tom~B. Brown, Miljan Martic, Shane Legg, and
  Dario Amodei.
\newblock Deep reinforcement learning from human preferences.
\newblock In \emph{Advances in Neural Information Processing Systems}, 2017.

\bibitem[Debreu(1954)]{debreu1954representation}
Gerard Debreu.
\newblock Representation of a preference ordering by a numerical function.
\newblock \emph{Decision Processes}, 3:\penalty0 159--165, 1954.

\bibitem[Dewey(2014)]{dewey2014reinforcement}
Daniel Dewey.
\newblock Reinforcement learning and the reward engineering principle.
\newblock In \emph{Proceedings of the AAAI Spring Symposium Series}, 2014.

\bibitem[Everitt et~al.(2017)Everitt, Krakovna, Orseau, Hutter, and
  Legg]{everitt2017reinforcement}
Tom Everitt, Victoria Krakovna, Laurent Orseau, Marcus Hutter, and Shane Legg.
\newblock Reinforcement learning with a corrupted reward channel.
\newblock In \emph{Proceedings of the International Joint Conference on
  Artificial Intelligence}, 2017.

\bibitem[Fedus et~al.(2019)Fedus, Gelada, Bengio, Bellemare, and
  Larochelle]{fedus2019hyperbolic}
William Fedus, Carles Gelada, Yoshua Bengio, Marc~G. Bellemare, and Hugo
  Larochelle.
\newblock Hyperbolic discounting and learning over multiple horizons.
\newblock \emph{arXiv preprint arXiv:1902.06865}, 2019.

\bibitem[Friston(2010)]{friston2010free}
Karl~J. Friston.
\newblock The free-energy principle: a unified brain theory?
\newblock \emph{Nature reviews neuroscience}, 11\penalty0 (2):\penalty0
  127--138, 2010.

\bibitem[Friston et~al.(2009)Friston, Daunizeau, and
  Kiebel]{friston2009reinforcement}
Karl~J. Friston, Jean Daunizeau, and Stefan~J. Kiebel.
\newblock Reinforcement learning or active inference?
\newblock \emph{PloS One}, 4\penalty0 (7):\penalty0 e6421, 2009.

\bibitem[Hadfield-Menell et~al.(2016)Hadfield-Menell, Dragan, Abbeel, and
  Russell]{hadfield2016cooperative}
Dylan Hadfield-Menell, Anca Dragan, Pieter Abbeel, and Stuart Russell.
\newblock Cooperative inverse reinforcement learning.
\newblock In \emph{Advances in Neural Information Processing Systems}, 2016.

\bibitem[Hadfield-Menell et~al.(2017{\natexlab{a}})Hadfield-Menell, Dragan,
  Abbeel, and Russell]{hadfield2016off}
Dylan Hadfield-Menell, Anca Dragan, Pieter Abbeel, and Stuart Russell.
\newblock The off-switch game.
\newblock In \emph{Proceedings of the International Joint Conference on
  Artificial Intelligence}, 2017{\natexlab{a}}.

\bibitem[Hadfield-Menell et~al.(2017{\natexlab{b}})Hadfield-Menell, Milli,
  Abbeel, Russell, and Dragan]{hadfield2017inverse}
Dylan Hadfield-Menell, Smitha Milli, Pieter Abbeel, Stuart Russell, and Anca
  Dragan.
\newblock Inverse reward design.
\newblock In \emph{Advances in Neural Information Processing Systems},
  2017{\natexlab{b}}.

\bibitem[Hafner et~al.(2020)Hafner, Ortega, Ba, Parr, Friston, and
  Heess]{hafner2020action}
Danijar Hafner, Pedro~A. Ortega, Jimmy Ba, Thomas Parr, Karl~J. Friston, and
  Nicolas Heess.
\newblock Action and perception as divergence minimization.
\newblock \emph{arXiv preprint arXiv:2009.01791}, 2020.

\bibitem[Hammond et~al.(2021)Hammond, Abate, Gutierrez, and
  Wooldridge]{hammond2021multi}
Lewis Hammond, Alessandro Abate, Julian Gutierrez, and Michael Wooldridge.
\newblock Multi-agent reinforcement learning with temporal logic
  specifications.
\newblock In \emph{Proceedings of the International Conference on Autonomous
  Agents and Multiagent Systems}, 2021.

\bibitem[Icarte et~al.(2018)Icarte, Klassen, Valenzano, and
  McIlraith]{icarte2018using}
Rodrigo~Toro Icarte, Toryn Klassen, Richard Valenzano, and Sheila McIlraith.
\newblock Using reward machines for high-level task specification and
  decomposition in reinforcement learning.
\newblock In \emph{Proceedings of the International Conference on Machine
  Learning}, 2018.

\bibitem[Jeon et~al.(2020)Jeon, Milli, and Dragan]{jeon2020reward}
Hong~Jun Jeon, Smitha Milli, and Anca Dragan.
\newblock Reward-rational (implicit) choice: A unifying formalism for reward
  learning.
\newblock In \emph{Advances in Neural Information Processing Systems}, 2020.

\bibitem[Jothimurugan et~al.(2020)Jothimurugan, Alur, and
  Bastani]{jothimurugan2020composable}
Kishor Jothimurugan, Rajeev Alur, and Osbert Bastani.
\newblock A composable specification language for reinforcement learning tasks.
\newblock In \emph{Advances in Neural Information Processing Systems}, 2020.

\bibitem[Karmarkar(1984)]{karmarkar1984new}
Narendra Karmarkar.
\newblock A new polynomial-time algorithm for linear programming.
\newblock In \emph{Proceedings of the Annual ACM Symposium on Theory of
  Computing}, 1984.

\bibitem[Knox and Stone(2009)]{knox2009interactively}
W.~Bradley Knox and Peter Stone.
\newblock Interactively shaping agents via human reinforcement: The {TAMER}
  framework.
\newblock In \emph{Proceedings of the International Conference on Knowledge
  Capture}, 2009.

\bibitem[Koopmans(1960)]{koopmans1960stationary}
Tjalling~C. Koopmans.
\newblock Stationary ordinal utility and impatience.
\newblock \emph{Econometrica: Journal of the Econometric Society}, pages
  287--309, 1960.

\bibitem[Kreps(1988)]{kreps1988notes}
David Kreps.
\newblock \emph{Notes on the Theory of Choice}.
\newblock Westview Press, 1988.

\bibitem[Krishna~Gottipati et~al.(2020)Krishna~Gottipati, Pathak, Nuttall,
  Chunduru, Touati, Ganapathi~Subramanian, Taylor, and
  Chandar]{krishna2020maximum}
Sai Krishna~Gottipati, Yashaswi Pathak, Rohan Nuttall, Raviteja Chunduru, Ahmed
  Touati, Sriram Ganapathi~Subramanian, Matthew~E. Taylor, and Sarath Chandar.
\newblock Maximum reward formulation in reinforcement learning.
\newblock In \emph{NeurIPS Workshop on Deep Reinforcement Learning}, 2020.

\bibitem[Kumar et~al.(2020)Kumar, Uesato, Ngo, Everitt, Krakovna, and
  Legg]{kumar2020realab}
Ramana Kumar, Jonathan Uesato, Richard Ngo, Tom Everitt, Victoria Krakovna, and
  Shane Legg.
\newblock {REALab}: An embedded perspective on tampering.
\newblock \emph{arXiv preprint arXiv:2011.08820}, 2020.

\bibitem[Li et~al.(2017)Li, Vasile, and Belta]{li2017reinforcement}
Xiao Li, Cristian-Ioan Vasile, and Calin Belta.
\newblock Reinforcement learning with temporal logic rewards.
\newblock In \emph{Proceedings of the International Conference on Intelligent
  Robots and Systems}, 2017.

\bibitem[Littman(2017)]{littmanRH}
Michael~L. Littman.
\newblock The reward hypothesis, 2017.
\newblock URL
  \url{https://www.coursera.org/lecture/fundamentals-of-reinforcement-learning/michael-littman-the-reward-hypothesis-q6x0e}.

\bibitem[Littman et~al.(2017)Littman, Topcu, Fu, Isbell, Wen, and
  MacGlashan]{littman2017environment}
Michael~L. Littman, Ufuk Topcu, Jie Fu, Charles Isbell, Min Wen, and James
  MacGlashan.
\newblock Environment-independent task specifications via {GLTL}.
\newblock \emph{arXiv preprint arXiv:1704.04341}, 2017.

\bibitem[MacGlashan et~al.(2015)MacGlashan, Babes-Vroman, desJardins, Littman,
  Muresan, Squire, Tellex, Arumugam, and Yang]{macglashan15}
James MacGlashan, Monica Babes-Vroman, Marie desJardins, Michael~L. Littman,
  Smaranda Muresan, Shawn Squire, Stefanie Tellex, Dilip Arumugam, and Lei
  Yang.
\newblock Grounding {E}nglish commands to reward functions.
\newblock In \emph{Proceedings of Robotics: Science and Systems}, 2015.

\bibitem[MacGlashan et~al.(2016)MacGlashan, Littman, Roberts, Loftin, Peng, and
  Taylor]{macglashan2016convergent}
James MacGlashan, Michael~L. Littman, David~L. Roberts, Robert Loftin, Bei
  Peng, and Matthew~E. Taylor.
\newblock Convergent actor critic by humans.
\newblock In \emph{Proceedings of the International Conference on Intelligent
  Robots and Systems}, 2016.

\bibitem[MacGlashan et~al.(2017)MacGlashan, Ho, Loftin, Peng, Wang, Roberts,
  Taylor, and Littman]{macglashan2017interactive}
James MacGlashan, Mark~K. Ho, Robert Loftin, Bei Peng, Guan Wang, David~L.
  Roberts, Matthew~E. Taylor, and Michael~L. Littman.
\newblock Interactive learning from policy-dependent human feedback.
\newblock In \emph{Proceedings of the International Conference on Machine
  Learning}. PMLR, 2017.

\bibitem[Mataric(1994)]{mataric1994reward}
Maja~J. Mataric.
\newblock Reward functions for accelerated learning.
\newblock In \emph{Proceedings of the International Conference on Machine
  Learning}, 1994.

\bibitem[Mitten(1974)]{mitten1974preference}
L.~G. Mitten.
\newblock Preference order dynamic programming.
\newblock \emph{Management Science}, 21\penalty0 (1):\penalty0 43--46, 1974.

\bibitem[Ng et~al.(1999)Ng, Harada, and Russell]{ng1999policy}
Andrew~Y. Ng, Daishi Harada, and Stuart Russell.
\newblock Policy invariance under reward transformations: Theory and
  application to reward shaping.
\newblock In \emph{Proceedings of the International Conference on Machine
  Learning}, 1999.

\bibitem[Ng et~al.(2000)Ng, Russell, et~al.]{ng2000algorithms}
Andrew~Y. Ng, Stuart~J. Russell, et~al.
\newblock Algorithms for inverse reinforcement learning.
\newblock In \emph{Proceedings of the International Conference on Machine
  Learning}, 2000.

\bibitem[Novoseller et~al.(2020)Novoseller, Wei, Sui, Yue, and
  Burdick]{novoseller2020dueling}
Ellen Novoseller, Yibing Wei, Yanan Sui, Yisong Yue, and Joel Burdick.
\newblock Dueling posterior sampling for preference-based reinforcement
  learning.
\newblock In \emph{Proceedings of the Conference on Uncertainty in Artificial
  Intelligence}, 2020.

\bibitem[Ortega et~al.(2018)Ortega, Maini, and the DeepMind
  Safety~Team]{ortega2018}
Pedro~A. Ortega, Vishal Maini, and the DeepMind Safety~Team.
\newblock Building safe artificial intelligence: specification, robustness, and
  assurance, 2018.
\newblock URL
  \url{https://medium.com/@deepmindsafetyresearch/building-safe-artificial-intelligence-52f5f75058f1}.

\bibitem[Pitis(2019)]{pitis2019rethinking}
Silviu Pitis.
\newblock Rethinking the discount factor in reinforcement learning: A decision
  theoretic approach.
\newblock In \emph{Proceedings of the AAAI Conference on Artificial
  Intelligence}, 2019.

\bibitem[Puterman(2014)]{puterman2014markov}
Martin~L. Puterman.
\newblock \emph{Markov Decision Processes: Discrete Stochastic Dynamic
  Programming}.
\newblock John Wiley \& Sons, 2014.

\bibitem[Russell and Norvig(1994)]{russell94}
Stuart~J. Russell and Peter Norvig.
\newblock \emph{Artificial Intelligence: {A} Modern Approach}.
\newblock Prentice-Hall, Englewood Cliffs, NJ, 1994.
\newblock ISBN 0-13-103805-2.

\bibitem[Shah et~al.(2021)Shah, Freire, Alex, Freedman, Krasheninnikov, Chan,
  Dennis, Abbeel, Dragan, and Russell]{shah2021benefits}
Rohin Shah, Pedro Freire, Neel Alex, Rachel Freedman, Dmitrii Krasheninnikov,
  Lawrence Chan, Michael~D. Dennis, Pieter Abbeel, Anca Dragan, and Stuart
  Russell.
\newblock Benefits of assistance over reward learning, 2021.
\newblock URL \url{https://openreview.net/forum?id=DFIoGDZejIB}.

\bibitem[Silver et~al.(2021)Silver, Singh, Precup, and
  Sutton]{silver2021reward}
David Silver, Satinder Singh, Doina Precup, and Richard~S. Sutton.
\newblock Reward is enough.
\newblock \emph{Artificial Intelligence}, page 103535, 2021.

\bibitem[Singh et~al.(2005)Singh, Barto, and Chentanez]{singh2005intrinsically}
Satinder Singh, Andrew~G. Barto, and Nuttapong Chentanez.
\newblock Intrinsically motivated reinforcement learning.
\newblock Technical report, University of Massachusetts at Amherst Department
  of Computer Science, 2005.

\bibitem[Singh et~al.(2009)Singh, Lewis, and Barto]{singh2009rewards}
Satinder Singh, Richard~L Lewis, and Andrew~G Barto.
\newblock Where do rewards come from?
\newblock In \emph{Proceedings of the Annual Conference of the Cognitive
  Science Society}, 2009.

\bibitem[Singh et~al.(2010)Singh, Lewis, Sorg, Barto, and
  Helou]{singh2010separating}
Satinder Singh, Richard~L. Lewis, Jonathan Sorg, Andrew~G. Barto, and Akram
  Helou.
\newblock On separating agent designer goals from agent goals: Breaking the
  preferences--parameters confound, 2010.

\bibitem[Sobel(1975)]{sobel1975ordinal}
Matthew~J. Sobel.
\newblock Ordinal dynamic programming.
\newblock \emph{Management science}, 21\penalty0 (9):\penalty0 967--975, 1975.

\bibitem[Sobel(2013)]{sobel2013discounting}
Matthew~J. Sobel.
\newblock Discounting axioms imply risk neutrality.
\newblock \emph{Annals of Operations Research}, 208\penalty0 (1):\penalty0
  417--432, 2013.

\bibitem[Sorg(2011)]{sorg2011optimal}
Jonathan Sorg.
\newblock \emph{The Optimal Reward Problem: Designing Effective Reward for
  Bounded Agents}.
\newblock PhD thesis, University of Michigan, 2011.

\bibitem[Sorg et~al.(2010)Sorg, Lewis, and Singh]{sorg2010reward}
Jonathan Sorg, Richard~L. Lewis, and Satinder Singh.
\newblock Reward design via online gradient ascent.
\newblock \emph{Advances in Neural Information Processing Systems}, 2010.

\bibitem[Sunehag and Hutter(2011)]{sunehag2011axioms}
Peter Sunehag and Marcus Hutter.
\newblock Axioms for rational reinforcement learning.
\newblock In \emph{Proceedings of the International Conference on Algorithmic
  Learning Theory}, 2011.

\bibitem[Sutton(2004)]{suttonwebRLhypothesis}
Richard~S. Sutton.
\newblock The reward hypothesis, 2004.
\newblock URL
  \url{http://incompleteideas.net/rlai.cs.ualberta.ca/RLAI/rewardhypothesis.html}.

\bibitem[Sutton and Barto(2018)]{sutton2018reinforcement}
Richard~S. Sutton and Andrew~G. Barto.
\newblock \emph{Reinforcement Learning: An Introduction}.
\newblock MIT Press, 2018.

\bibitem[Syed et~al.(2008)Syed, Bowling, and Schapire]{syed2008apprenticeship}
Umar Syed, Michael Bowling, and Robert~E. Schapire.
\newblock Apprenticeship learning using linear programming.
\newblock In \emph{Proceedings of the International Conference on Machine
  Learning}, 2008.

\bibitem[Szepesvári(2020)]{csabaRLhypothesis}
Csaba Szepesvári.
\newblock Constrained {MDP}s and the reward hypothesis, 2020.
\newblock URL
  \url{http://readingsml.blogspot.com/2020/03/constrained-mdps-and-reward-hypothesis.html}.

\bibitem[Tasse et~al.(2020)Tasse, James, and Rosman]{tasse2020boolean}
Geraud~Nangue Tasse, Steven James, and Benjamin Rosman.
\newblock A {B}oolean task algebra for reinforcement learning.
\newblock In \emph{Advances in Neural Information Processing Systems}, 2020.

\bibitem[Toro~Icarte et~al.(2018)Toro~Icarte, Klassen, Valenzano, and
  McIlraith]{toro2018teaching}
Rodrigo Toro~Icarte, Toryn~Q. Klassen, Richard Valenzano, and Sheila~A.
  McIlraith.
\newblock Teaching multiple tasks to an {RL} agent using {LTL}.
\newblock In \emph{Proceedings of the International Conference on Autonomous
  Agents and Multiagent Systems}, 2018.

\bibitem[von Neumann and Morgenstern(1953)]{vonneumann1953theory}
John von Neumann and Oskar Morgenstern.
\newblock \emph{Theory of Games and Economic Behavior}.
\newblock Princeton University Press, 1953.

\bibitem[Weng(2011)]{weng2011markov}
Paul Weng.
\newblock Markov decision processes with ordinal rewards: Reference point-based
  preferences.
\newblock In \emph{Proceedings of the International Conference on Automated
  Planning and Scheduling}, 2011.

\bibitem[White(2017)]{white2017unifying}
Martha White.
\newblock Unifying task specification in reinforcement learning.
\newblock In \emph{Proceedings of the International Conference on Machine
  Learning}, 2017.

\bibitem[Williams et~al.(2018)Williams, Gopalan, Rhee, and
  Tellex]{williams2018learning}
Edward~C. Williams, Nakul Gopalan, Mine Rhee, and Stefanie Tellex.
\newblock Learning to parse natural language to grounded reward functions with
  weak supervision.
\newblock In \emph{Proceedings of the International Conference on Robotics and
  Automation}, 2018.

\bibitem[Wilson et~al.(2012)Wilson, Fern, and Tadepalli]{wilson2012bayesian}
Aaron Wilson, Alan Fern, and Prasad Tadepalli.
\newblock A {B}ayesian approach for policy learning from trajectory preference
  queries.
\newblock In \emph{Advances in Neural Information Processing Systems}, 2012.

\bibitem[Wirth et~al.(2017)Wirth, Akrour, Neumann, and
  F{\"u}rnkranz]{wirth2017survey}
Christian Wirth, Riad Akrour, Gerhard Neumann, and Johannes F{\"u}rnkranz.
\newblock A survey of preference-based reinforcement learning methods.
\newblock \emph{The Journal of Machine Learning Research}, 18\penalty0
  (1):\penalty0 4945--4990, 2017.

\bibitem[Xu et~al.(2020)Xu, Wang, Yang, Singh, and Dubrawski]{xu2020preference}
Yichong Xu, Ruosong Wang, Lin Yang, Aarti Singh, and Artur Dubrawski.
\newblock Preference-based reinforcement learning with finite-time guarantees.
\newblock \emph{Advances in Neural Information Processing Systems}, 33, 2020.

\bibitem[Zheng et~al.(2020)Zheng, Oh, Hessel, Xu, Kroiss, van Hasselt, Silver,
  and Singh]{zheng2020can}
Zeyu Zheng, Junhyuk Oh, Matteo Hessel, Zhongwen Xu, Manuel Kroiss, Hado van
  Hasselt, David Silver, and Satinder Singh.
\newblock What can learned intrinsic rewards capture?
\newblock In \emph{Proceedings of the International Conference on Machine
  Learning}, 2020.

\end{thebibliography}
